\documentclass[preprint,10pt]{elsarticle}
\usepackage{fullpage}

\usepackage{amsfonts,amssymb,amsbsy,amsmath,mathtools,multirow,cases}
\usepackage{amsmath, scalerel}

\usepackage{graphicx}
\usepackage{pgfplots}
\usepackage{amsthm}
\usepackage{graphicx,epstopdf}

\usepackage{stmaryrd}
\theoremstyle{remark}

\DeclareMathAlphabet\mathbfcal{OMS}{cmsy}{b}{n}

\usepgfplotslibrary{external}  
\usepackage{calligra,calrsfs,mathrsfs}
\usepackage{ulem,cancel}
\usepackage{subcaption}
\usepackage{tkz-euclide}
\usetikzlibrary[patterns]
\usepackage[ruled,vlined]{algorithm2e}
\usepackage{algpseudocode}
\usepackage{dsfont}
\usepackage{caption}
\usepackage{epstopdf}
\usepackage{epsfig}
\usepackage{natbib}
\usepackage{grffile}
\usepackage{tabularx}
\usepackage{amsopn}
\usepackage{enumerate}
\usepackage{amsfonts}
\usepackage{hyperref}
\usepackage{cleveref}
\usepackage{changes}

\renewcommand{\[}{\left[}
\renewcommand{\]}{\right]}

\newcommand{\vvvert}{|\kern-1pt|\kern-1pt|}

\newcommand{\bd}{\textbf{d}}

\newcommand{\by}{\textbf{y}}

\newcommand{\bY}{\mathbf{Y}}

\newcommand{\btheta}{\boldsymbol{\theta}}

\newcommand{\EE}{\mathbb{E}}

\newcommand{\RR}{\mathbb{R}}

\newcommand{\CN}{\mathcal{N}}

\newcommand{\argmax}{\operatornamewithlimits{argmax}}

\newcommand{\DKL}{D_{\mathrm{KL}}}

\newtheorem{prop}{Proposition}

\crefname{section}{Sec.}{Sec.}
\Crefname{section}{Section}{Sections}
\crefname{subsection}{Sec.}{Sec.}
\Crefname{subsection}{Section}{Sections}
\crefname{figure}{Fig.}{Fig.}
\Crefname{figure}{Figure}{Figures}
\crefname{equation}{Eqn.}{Eqn.}
\Crefname{equation}{Equation}{Equations}
\crefname{table}{Table}{Tables}
\Crefname{table}{Table}{Tables}

\crefname{prop}{Proposition}{Propositions}
\Crefname{prop}{Proposition}{Propositions}

\crefdefaultlabelformat{#2\textup{#1}#3}

\usepackage[version=4]{mhchem}

\usepackage{longtable,tabularx}
\SetKwInput{KwInput}{Input}                %
\SetKwInput{KwOutput}{Output}              %

\usepackage{diagbox}
\usepackage{booktabs}
\usepackage{float}

\usepackage{eucal}
\usepackage{amsmath}

\newdefinition{rem}{Remark}

\allowdisplaybreaks

\usepackage{multirow}

\newcolumntype{G}{>{\centering\columncolor{gray!20!white}}p{0.2\textwidth}}
\newcolumntype{C}{>{\centering\arraybackslash}p{0.2\textwidth}}

\begin{document}

\begin{frontmatter}

\title{Variational Bayesian Optimal Experimental Design with Normalizing Flows}

\cortext[cor1]{~Corresponding author}
\address[a]{Department of Mechanical Engineering, University of Michigan, Ann Arbor, MI, 48109, USA}
\address[b]{Department of Aerospace Engineering, University of Michigan, Ann Arbor, MI, 48109, USA}
\address[d]{Ford Research \& Innovation Center, Dearborn, MI, 48121, USA}
\address[c]{The Dow Chemical Company, Core R\&D, Engineering and Process Science, Lake Jackson, TX 77566, USA}
\author[a]{Jiayuan~Dong\corref{cor1}}
\ead{jiayuand@umich.edu}
\author[b]{Christian~Jacobsen}
\ead{csjacobs@umich.edu}
\author[a,c]{Mehdi~Khalloufi}
\ead{khalloufi.mehdi@gmail.com}
\author[d]{Maryam~Akram}
\ead{makram13@ford.com}
\author[d]{Wanjiao~Liu}
\ead{lwanjiao@ford.com}
\author[b]{Karthik~Duraisamy}
\ead{kdur@umich.edu}
\author[a]{Xun~Huan}
\ead{xhuan@umich.edu}

\begin{abstract}
Bayesian optimal experimental design (OED) seeks experiments that maximize the expected information gain (EIG) in model parameters. Directly estimating the EIG using nested Monte Carlo is computationally expensive and requires an explicit likelihood. Variational OED (vOED), in contrast, estimates a lower bound of the EIG without likelihood evaluations by approximating the posterior distributions with variational forms, and then tightens the bound by optimizing its variational parameters. We  introduce the use of normalizing flows (NFs) for representing variational distributions in vOED; we call this approach vOED-NFs. 
Specifically, we adopt NFs with a conditional invertible neural network architecture built from compositions of coupling layers, and enhanced with a summary network for data dimension reduction. 
We present Monte Carlo estimators to the lower bound along with gradient expressions to enable a gradient-based simultaneous optimization of the variational parameters and the design variables. 
The vOED-NFs algorithm is then validated in two benchmark problems, and demonstrated on a partial differential equation-governed application of cathodic electrophoretic deposition and an implicit likelihood case with stochastic modeling of aphid population. The findings suggest that a composition of 4--5 coupling layers is able to achieve lower EIG estimation bias, under a fixed budget of forward model runs, compared to previous approaches. The resulting NFs produce approximate posteriors that agree well with the true posteriors, able to capture non-Gaussian and multi-modal features effectively. 
\end{abstract}


\end{frontmatter}

\section{Introduction}
\label{sec: intro}

Mathematical models are central for understanding and making predictions in physical systems. Parameters in models are often uncertain, but their uncertainty can be reduced when information is gained from experimental observations. Since experiments are costly in many applications, optimal experimental design (OED) (see, e.g., \cite{Pukelsheim_06_SIAM, Chaloner_95_Review, Atkinson07, Ryan_16_Review, Rainforth_23_Review, Huan_24_Optimal}) can bring substantial resource savings by identifying the experiments that produce the most valuable data. 

Bayesian OED further incorporates the prior and posterior uncertainty in its objective function. The seminal paper by Lindley \cite{Lindley56} proposed to use mutual information (MI) between model parameters and experimental data as the design criterion, which is equivalent to the expected information gain (EIG) in the model parameters.
The EIG, however, is generally intractable to compute and has to be approximated numerically. 
One strategy entails directly approximating the EIG. For example, Ryan~\cite{Ryan2003} introduced a nested Monte Carlo (NMC) estimator for the EIG, but it is computationally expensive with the number of likelihood evaluations scaling as the \emph{product} of sample sizes of the nested Monte Carlo loops.
Advanced numerics have been developed to accelerate NMC computations through surrogate modeling~\cite{Huan_13_Simulation, Huan_14_Gradient}, reusing samples across nested loops~\cite{Huan_13_Simulation}, and
`Gaussianizing' the posterior via Laplace approximations~\cite{long2013, Yu2018, keyi23, beck2018fast}. 
Another estimator similar to the NMC is the prior contrastive estimator (PCE) \cite{Foster_20_Unified}.
A key difference is that PCE reuses each outer loop sample once in the inner loop, resulting the estimator mean to be negatively biased (i.e., a lower bound of the EIG). 
The above techniques, however, all require the ability to evaluate the likelihood function and cannot accommodate implicit or intractable likelihood settings that often arise from stochastic models.
An EIG lower bound that is able to handle intractable likelihood is the LB-KLD~\cite{Ziqiao_20_LBKLD}, which can be derived from Shannon’s entropy power inequality. 

Another strategy is to adopt a \emph{variational} bound for the EIG~\cite{Poole_19_Bounds} and tighten the bound by optimizing its variational parameters. (Note that PCE and LBKLD do not have variational parameters for tightening the bounds, and therefore are not considered to be variational bounds.) We call these \emph{variational OED} (vOED) approaches.  
For example, the tractable unnormalized Barber--Agakov (TUBA) lower bound~\cite{Poole_19_Bounds} incorporates the tuning of a `critic' function, and the Nguyen--Wainwright--Jordan (NWJ) bound \cite{Nguyen_10_NWJ}, also known as the mutual information neural estimation f-divergence (MINE-f) bound \cite{Belghazi_18_Mine}, can be shown to be a special case of TUBA and has been used by Kleinegesse and Gutmann~\cite{Kleinegesse_20_MINEBED} in the OED context. 
Another lower bound, the information noise-contrastive estimation (InfoNCE) bound \cite{Oord_18_InfoNCE}, takes a similar form as PCE, but replaces the likelihood with an exponentiated critic function.
More recently, Ivanova \textit{et al.} \cite{Desi_21_Idad} extended the NWJ and InfoNCE bounds to the sequential OED setting. All of these bounds can accommodate implicit likelihoods.

A particular type of variational bounds, known as the Barber--Agakov (BA) lower bound~\cite{BarberAgakov03}, emerges from approximating the posterior density function directly. A potential advantage of the BA bound is its ability to enable both density evaluation and sampling from the resulting approximate posteriors. This is in contrast to several other lower bound choices, such as the NWJ bound that allows density evaluation but not sampling, and the InfoNCE bound that does not offer density or sampling.
Foster \textit{et al.} \cite{Foster_19_VBOED} were the first to use the BA bound in the OED setting, but remained with relatively simple variational distributions such as Gaussian, Bernoulli, and Gamma. 
The approximations thus can deteriorate %
when the true posterior distributions depart from these forms. 
Improving upon their work, Foster \textit{et al.} \cite{Foster_20_Unified} then introduced the adaptive contrastive estimation (ACE) lower bound based on the PCE but used a variational biasing distribution for the inner loop sampling. However, ACE required explicit likelihood.    

One of the key contributions of this paper is therefore to reduce the BA bound bias in estimating the EIG by improving the accuracy of posterior approximation through enriching the space of variational distributions using normalizing flows (NFs)~\cite{Dinh_16_NVP, Rezende_16_Variational, Papamakarios_21_NFsReview,Kobyzev_20_NFsReview}. We achieve this while maintaining the ability to handle implicit likelihoods.

{
NFs are rooted in measure transport theory~\cite{Villani2008,Marzouk_16_Transport,Spantini2018} which seek an invertible mapping between a target distribution (e.g., the posterior) and a reference distribution (e.g., a standard normal). Once such a transport map is established, the density function of the target distribution can be obtained from the density of the reference, and vice versa, through the standard change-of-variable formula. Hence, an approximate posterior can be represented by a parameterized transport map, and the problem becomes finding the best transport map from its parametric class. In vOED, `best' is meant by maximizing the lower bound; in the related field of expectation propagation~\cite{Minka2001} (a relative of variational inference but uses the `reverse' KL divergence), `best' is meant by minimizing the Kullback--Lebiler (KL) divergence from the approximate posterior to the true posterior---we will show that these two notions are equivalent in expectation. 
}

{
Different structures and parameterizations of transport maps have been proposed. Triangular and block-triangular forms~\cite{Bogachev_05_Triangular} are often parameterized using multivariate polynomials~\cite{ElMoselhy2012, Marzouk_16_Transport}, radial basis functions~\cite{Marzouk_16_Transport},  tensor decomposition~\cite{Cui_21_tensor}, and partially convex potential maps and conditional optimal transport flows~\cite{Wang_23_COT}.
Invertibility is then enforced through local monotonicity constraints, or setting parameterizations in ways to guarantee monotonicity \cite{Marzouk_16_Transport, Wang_23_COT}. 
For instance, Baptista \textit{et al.} \cite{Baptista_23_Representation} proposed a rectification operator that can always transform sufficiently smooth non-monotone functions into monotone component functions of a triangular map, and also demonstrated the elimination of spurious local minima that often arises in the training of neural network-based maps. 
Furthermore, the triangular structures provide a convenient means to extract conditional densities (e.g., the posterior or likelihood) and to compute the Jacobian of transformation. 

}

{
Elsewhere, efforts in the machine learning community in creating practical transport maps focused on composing together invertible, often neural network-based, functions, known as normalizing flows. For recent reviews of NFs, see~\cite{Kobyzev_20_NFsReview, Papamakarios_21_NFsReview,Papamakarios_21_NFsReview}. 
While first represented as a simple 3-trainable-parameter function~\cite{Tabak_10_Density} and then as a composition of localized radial expansion to achieve greater expressiveness~\cite{Tabak_13_Flow}, the architectural form of NFs has evolved rapidly. 
Examples include planar and radial flows \cite{Rezende_16_Variational,Tomczak_16_Householder, louizos_17_MultiplicativeNF, Berg_19_Sylvester}, coupling layers and autoregressive flows \cite{Dinh_14_Nice, Dinh_16_NVP, kingma_16_IAF, Papamakarios_17_MAF, Kingma_18_Glow}, splines \cite{durkan_19_Spline}, residual flows \cite{Chen_19_Residual} and neural ordinary differential equations \cite{Chen_18_NODE}. While NFs have been used for approximating the posterior in a variational inference context (see \cite{Rezende_16_Variational} and several subsequent works), their application in OED  have been only lightly explored. Kennamer \textit{et al.} \cite{Kennamer_Walton_Ihler_2023} demonstrated the use of NFs for EIG estimation focusing on generalized linear models (GLMs),
while Orozco \textit{et al.} \cite{Orozco_24_Probabilistic} applied NFs to OED for image datasets and studied posterior samples of the test error---neither has yet made comparisons with alternative OED estimators, such as NMC.

In this paper, we propose and investigate the use of NFs for approximating posterior distributions \textit{en route} to optimizing the BA bound estimates of the EIG for vOED; we call this approach \textbf{vOED-NFs}. 
The key novelty and contributions of this paper are as follows.
\begin{itemize}
    \item We present the use of NFs for estimating the EIG lower bound in vOED, and illustrate its sampling efficiency (lower bias) compared to previous approaches.
    \item We show the ability of NFs trained for vOED to achieve good posterior approximations, including non-Gaussian and multi-modal distributions. 
    \item We validate vOED-NFs against established reference methods, and demonstrate vOED-NFs on a partial differential equation (PDE)-governed application of cathodic electrophoretic deposition. 
    \item We illustrate vOED-NFs in handling an implicit likelihood case with stochastic modeling of aphid population.
\end{itemize}
 
This paper is structured as follows. \Cref{section: ProblemFormulation} reviews the OED and vOED problem formulations. \Cref{section: ComputationalMethods} illustrates the computational methods for solving the vOED problem. 
\Cref{section: Applications} then presents a number of numerical experiments, starting with simple benchmarks to validate vOED-NFs against existing established approaches, and then demonstrating it on a PDE-governed design problem for cathodic electrophoretic deposition (e-coating) studied in the automotive industry and an implicit case for aphid population modeling. 
The paper ends with conclusions and future work in \cref{s: Conclusions}.

\section{Problem Formulation}
\label{section: ProblemFormulation}

\subsection{Bayesian optimal experimental design}
\label{subsection: BOED}

We adopt the following notation: upper case for random variable, lower case for realization, bold for vector or matrix, and subscript in a probability density function (PDF) is generally omitted but retained in some cases for clarification or emphasis; for example, $p_{\mathbf{X}}(\mathbf{X}=\mathbf{x}) = p(\textbf{x})$ denotes the PDF of random vector $\mathbf{X}$ evaluated at value $\mathbf{X}=\mathbf{x}$. 
When an experiment is conducted under design $\mathbf{d} \in \mathcal{D} \subseteq \mathbb{R}^{n_d}$ and yields an observation $\mathbf{Y} = \mathbf{y} \in \mathbb{R}^{n_y}$, 
the PDF for the model parameters $\boldsymbol{\Theta}\in \mathbb{R}^{n_{\theta}}$ is updated via Bayes' rule by
\begin{align}
p(\boldsymbol{\theta}|\mathbf{y}, \mathbf{d}) = \frac{p(\boldsymbol{\theta})\,p(\mathbf{y}|\boldsymbol{\theta},\mathbf{d})}{p(\mathbf{y}|\mathbf{d})},
\label{e:Bayes}
\end{align}
where  $p(\boldsymbol{\theta})$
is the prior PDF (we assume the prior does not depend on $\mathbf{d}$), $p(\mathbf{y}|\boldsymbol{\theta},\mathbf{d})$ is the data likelihood, $p(\mathbf{y}|\mathbf{d})$ is the marginal likelihood (model evidence), and $p(\boldsymbol{\theta}|\mathbf{y}, \mathbf{d})$ is the posterior PDF. 
The likelihood can be computed based on an underlying observation model, for example,
\begin{align}
    \mathbf{Y} = \mathbf{G}(\boldsymbol{\Theta}, \mathbf{d}) + \boldsymbol{\mathcal{E}},
\end{align}
where $\mathbf{G}:\mathbb{R}^{n_\theta} \times \mathbb{R}^{n_d} \rightarrow \mathbb{R}^{n_y}$ is the forward model (e.g., from solving a governing system of PDEs) and $\boldsymbol{\mathcal{E}}$ is the random noise associated with the observation model. Each likelihood evaluation thus entails computing $p(\mathbf{y}|\boldsymbol{\theta},\mathbf{d})=p_{\boldsymbol{\mathcal{E}}}(\mathbf{y}-\mathbf{G}(\boldsymbol{\theta},\mathbf{d}))$, which requires a forward model solve.

We adopt the EIG on $\boldsymbol{\Theta}$ to reflect the expected utility of performing an experiment. Mathematically, the EIG is the expected KL divergence from the parameter prior to the posterior, and can be interpreted as the expected parameter uncertainty reduction due to the experiment:
\begin{align}
    U(\mathbf{d}) 
    = \mathbb{E}_{\mathbf{Y}|\mathbf{d}} 
    \[  D_{\text{KL}}(\,p_{\boldsymbol{\Theta}| \mathbf{Y}, \mathbf{d}}\,||\,p_{\boldsymbol{\Theta}}\,)\]
    = \iint p(\boldsymbol{\theta}, \mathbf{y} | \mathbf{d})\ln \[\frac{p(\boldsymbol{\theta}| \mathbf{y}, \mathbf{d})}{p(\boldsymbol{\theta})}\] d\boldsymbol{\theta}\, d\mathbf{y}.
    \label{e:Ud}
\end{align}
The Bayesian OED problem then entails solving for the optimal design:
\begin{align}
    \mathbf{d}^{\ast} = \argmax_{\mathbf{d} \in \mathcal{D}} U(\mathbf{d}).
\end{align}

\subsection{Variational optimal experimental design}
\label{subsection: VBOED}

We summarize below the variational OED (vOED) approach introduced by Foster \textit{et al.}~\cite{Foster_19_VBOED}. vOED replaces the posterior or marginal likelihood with an approximate PDF $q(\cdot)$ parameterized by $\boldsymbol{\lambda}$ $\in \RR^{n_{\boldsymbol{\lambda}}}$. 
When approximating the posterior, the expected utility from \cref{e:Ud} becomes
\begin{align}
    U_L(\mathbf{d};\boldsymbol{\lambda}) = \mathbb{E}_{\boldsymbol{\Theta},\mathbf{Y}|\mathbf{d}} 
    \[ \ln \frac{q(\boldsymbol{\theta}| \mathbf{y};\boldsymbol{\lambda})}{p(\boldsymbol{\theta} )} \]  = \iint p(\boldsymbol{\theta}, \mathbf{y}| \mathbf{d})\ln \[\frac{q(\boldsymbol{\theta}| \mathbf{y};\boldsymbol{\lambda})}{p(\boldsymbol{\theta} )}\] d\boldsymbol{\theta}\,  d\mathbf{y}.
    \label{e:UL}
\end{align}
Note that the outer expectation is still taken with respect to the \emph{true} distribution $p(\boldsymbol{\theta}, \mathbf{y}| \mathbf{d})$.
$U_L(\mathbf{d}; \boldsymbol{\lambda})$ is known as the Barber--Agakov bound \cite{BarberAgakov03}; it is a lower bound for $U(\mathbf{d})$:
\begin{align}
    U(\mathbf{d}) - U_L(\mathbf{d};\boldsymbol{\lambda}) 
    = \mathbb{E}_{\mathbf{Y}|\mathbf{d}} 
    [  D_{\text{KL}}(\,p_{\boldsymbol{\Theta}| \mathbf{Y}, \mathbf{d}}\,||\,q_{\boldsymbol{\Theta}| \mathbf{Y};\boldsymbol{\lambda}} \,)]
    \geq 0.
    \label{e:UL_bound}
\end{align}
The bound is tight if and only if $q(\boldsymbol{\theta}|\mathbf{y};\boldsymbol{\lambda}) = p(\boldsymbol{\theta}|\mathbf{y}, \mathbf{d})$ for all $\boldsymbol{\theta}$ and $\mathbf{y}$. 
The best (tightest) lower bound at a given $\mathbf{d}$ is then
\begin{align}
    U_L(\mathbf{d};\boldsymbol{\lambda}^{\ast})  =  \max_{\boldsymbol{\lambda}} 
    U_L(\mathbf{d};\boldsymbol{\lambda}).
    \label{e:UL_opt}
\end{align}
We note that the ordering of posterior approximation quality for $q(\btheta|\by;\boldsymbol{\lambda})$ as measured by $U_L(\bd;\boldsymbol{\lambda})$ is preserved when measured by the expected KL divergence in the form of moment projection as used in expectation propagation~\cite{Minka2001} (i.e., the `reverse' KL divergence compared to the information projection form used in variational inference). Hence, the approximate posterior resulting from a tighter lower bound is also closer to the true posterior in this expected KL sense. We make this precise in the following proposition. 
\begin{prop}
\label{prop:ordering}
Consider probability densities $q(\btheta|\by;\boldsymbol{\lambda}_1)$ and $q(\btheta|\by;\boldsymbol{\lambda}_2)$ formed at variational parameter values $\boldsymbol{\lambda}_1$ and $\boldsymbol{\lambda}_2$, respectively, both as approximations to the true posterior density $p(\btheta|\by,\bd)$. Then, 
\begin{align}
U_L(\bd;\boldsymbol{\lambda}_1)\leq U_L(\bd;\boldsymbol{\lambda}_2)
\label{e:UL_ordering}
\end{align}
if and only if 
\begin{align}
\EE_{\bY|\bd}\left[\DKL(p_{\btheta|\by,\bd}\,||\,q_{\btheta|\by;\boldsymbol{\lambda}_1})\right] \geq \EE_{\bY|\bd}\left[\DKL(p_{\btheta|\by,\bd}\,||\,q_{\btheta|\by;\boldsymbol{\lambda}_2})\right].
\label{e:EKL_ordering}
\end{align}
\end{prop}
A proof of the proposition is provided in \ref{app:prop_proof}.

With $U_L(\mathbf{d};\boldsymbol{\lambda}^{\ast})$ being the best variational approximation to the true EIG at $\mathbf{d}$, the vOED problem looks for the design that maximizes the tightest bound by simultaneously optimizing for $\mathbf{d}$ and $\boldsymbol{\lambda}$:
\begin{align}
    \mathbf{d}^{\ast},\boldsymbol{\lambda}^{\ast} = \argmax_{\mathbf{d}\in\mathcal{D},\boldsymbol{\lambda}}
    U_L(\mathbf{d};\boldsymbol{\lambda}).
    \label{e:UL_OED}
\end{align} 
When $U_L(\mathbf{d};\boldsymbol{\lambda})$ is differentiated with respect to $\mathbf{d}$ and $\boldsymbol{\lambda}$, their gradients are
\begin{align}
    \nabla_{\boldsymbol{\lambda}} U_L(\mathbf{d}; \boldsymbol{\lambda}) 
    &= \mathbb{E}_{\boldsymbol{\Theta}} \{ \mathbb{E}_{\mathbf{Y}|\boldsymbol{\Theta}, \mathbf{d}} 
    [\nabla_{\boldsymbol{\lambda}}
    \ln q(\boldsymbol{\theta}|\mathbf{y}; \boldsymbol{\lambda})] \}, \label{e:UL_grad_lambda}\\
    \nonumber
    \nabla_{\mathbf{d}} U_L(\mathbf{d};\boldsymbol{\lambda}) &= \mathbb{E}_{\boldsymbol{\Theta}} \{ \nabla_{\mathbf{d}} \mathbb{E}_{\mathbf{Y}|\boldsymbol{\Theta}, \mathbf{d}} 
    [\ln q(\boldsymbol{\theta}|\mathbf{y}; \boldsymbol{\lambda}) ] \} = \mathbb{E}_{\boldsymbol{\Theta}} \left\{ \int [\nabla_{\mathbf{d}}p( \mathbf{y}|\boldsymbol{\theta}, \mathbf{d})] \ln q(\boldsymbol{\theta}|\mathbf{y}; \boldsymbol{\lambda})  \right\}  \\
    & = \mathbb{E}_{\boldsymbol{\Theta}} \{ \mathbb{E}_{\mathbf{Y}|\boldsymbol{\Theta}, \mathbf{d}} 
    [\ln q(\boldsymbol{\theta}|\mathbf{y}; \boldsymbol{\lambda}) \nabla_{\mathbf{d}} \ln p(\mathbf{y}|\boldsymbol{\theta}, \mathbf{d})  ] \}, \label{e:UL_grad_d}
\end{align}
where the last equality uses the identity $\nabla_{\mathbf{d}} \ln p(\mathbf{y}|\boldsymbol{\theta}, \mathbf{d}) = \frac{\nabla_{\mathbf{d}} p(\mathbf{y}|\boldsymbol{\theta}, \mathbf{d})}{p(\mathbf{y}|\boldsymbol{\theta}, \mathbf{d})}$. Access to these gradient formulas allows the adoption of gradient-based optimization algorithms to solve \cref{e:UL_OED}.

Alternatively, one can approximate the marginal likelihood instead of the posterior as introduced in \cite{Foster_19_VBOED}, and the expected utility becomes
\begin{align}
    U_U(\mathbf{d};\boldsymbol{\lambda}) = \mathbb{E}_{\boldsymbol{\Theta}, \mathbf{Y}| \mathbf{d}} 
    \[ \ln \frac{p(\mathbf{y}|\boldsymbol{\theta}, \mathbf{d})}{q(\mathbf{y}; \boldsymbol{\lambda})}\]  = \iint p(\boldsymbol{\theta}, \mathbf{y}| \mathbf{d}) \ln \[\frac{p(\mathbf{y}|\boldsymbol{\theta}, \mathbf{d})}{q(\mathbf{y}; \boldsymbol{\lambda})}\] d\boldsymbol{\theta}\, d\mathbf{y},
    \label{e:UU}
\end{align}
where the posterior-to-prior density ratio has been replaced with likelihood-to-marginal-likelihood density ratio via Bayes' rule in \cref{e:Bayes}.
Again, the outer expectation remains with respect to the \emph{true} distribution $p(\boldsymbol{\theta}, \mathbf{y}| \mathbf{d})$.
$U_U$ forms an upper bound for $U(\mathbf{d})$:
\begin{align}
    U_U(\mathbf{d};\boldsymbol{\lambda}) - U(\mathbf{d}) 
    = D_{\text{KL}}(\,p_{\mathbf{y}|\mathbf{d}} \,||\,q_{\mathbf{y};\boldsymbol{\lambda}} \,)
    \geq 0.
    \label{e:UU_bound}
\end{align}
The bound is tight if and only if $q(\mathbf{y}; \boldsymbol{\lambda}) = p(\mathbf{y}|\mathbf{d})$ for all $\mathbf{y}$.
The best (tightest) upper bound at a given $\mathbf{d}$ is then
\begin{align}
U_U(\mathbf{d};\boldsymbol{\lambda}^{\ast})  =  \min_{\boldsymbol{\lambda}} 
U_U(\mathbf{d};\boldsymbol{\lambda}).
\label{e:UU_opt}
\end{align}
The corresponding vOED problem using $U_U(\mathbf{d};\boldsymbol{\lambda})$ involves solving for
\begin{align}
    \mathbf{d}^{\ast} = \argmax_{\mathbf{d}\in\mathcal{D}}\min_{\boldsymbol{\lambda}}
    U_U(\mathbf{d};\boldsymbol{\lambda}).
    \label{e:UU_OED}
\end{align} 
When $U_U(\mathbf{d};\boldsymbol{\lambda})$ is differentiate with respect to $\boldsymbol{\lambda}$, its gradient is
\begin{align}
    \nabla_{\boldsymbol{\lambda}} U_U(\mathbf{d}; \boldsymbol{\lambda}) 
    = -\mathbb{E}_{\mathbf{Y}|\mathbf{d}} 
    [\nabla_{\boldsymbol{\lambda}}
    \ln q(\mathbf{y}; \boldsymbol{\lambda})].
    \label{e:UU_grad_lambda}
\end{align}
The use of $U_U(\mathbf{d};\boldsymbol{\lambda})$ can offer advantages when $n_y \ll n_{\theta}$ so that the variational density approximation is applied to a lower-dimensional space. However, the use of $U_U(\mathbf{d};\boldsymbol{\lambda})$ involves solving a more difficult maximin saddle-point problem in \cref{e:UU_OED}, and its two optimizations cannot be combined like in \cref{e:UL_OED}.
In this paper, we will primarily focus on the lower bound 
$U_L(\mathbf{d};\boldsymbol{\lambda})$ and its corresponding vOED problem in \cref{e:UL_OED}.

\section{Computational Methods}
\label{section: ComputationalMethods}

The expected utility in \cref{e:Ud} is generally intractable and needs be estimated numerically. One approach is to use the NMC estimator~\cite{Ryan2003}:
\begin{align}
     U(\mathbf{d}) 
     & \approx \widehat{U}_{\text{NMC}}(\mathbf{d}) \coloneqq
     \frac{1}{N_{\text{out}}} \sum_{i=1}^{N_{\text{out}}} \left\{ \ln p(\mathbf{y}^{(i)}|\boldsymbol{\theta}^{(i)}, \mathbf{d}) - \ln \[\frac{1}{N_{\text{in}}}\sum_{j=1}^{N_{\text{in}}} p(\mathbf{y}^{(i)}|\boldsymbol{\theta}^{(i,j)}, \mathbf{d})\] \right\},  %
     \label{e:MC}
\end{align}
where samples $\boldsymbol{\theta}^{(i)} \sim p(\boldsymbol{\theta})$, $\mathbf{y}^{(i)} \sim p(\mathbf{y}| \boldsymbol{\theta}^{(i)},\mathbf{d})$, and $\boldsymbol{\theta}^{(i, j)} \sim p(\boldsymbol{\theta})$ are drawn from the prior and likelihood. Calculating the NMC estimator $\widehat{U}_{\text{NMC}}$ requires evaluating the likelihood $p(\mathbf{y}| \boldsymbol{\theta},\mathbf{d})$ and cannot directly handle implicit likelihoods settings.
$\widehat{U}_{\text{NMC}}$ is a biased estimator to $U$ under any finite inner-loop sample size $N_{\text{in}}$, but it is asymptotically unbiased as $N_{\text{in}}\rightarrow \infty$.
Properties of the NMC estimator has been studied extensively~\cite{Ryan2003,Feng2019,beck2018fast, Rainforth_18_NMC}, and we will use it for reference comparison in this work when applicable. 

For vOED, the lower and upper bounds in \cref{e:UL,e:UU} require numerical approximations for their expectation operators, for example through standard Monte Carlo (MC) estimation:
\begin{align}
      U_L(\mathbf{d};\boldsymbol{\lambda}) &\approx \widehat{U}_L(\mathbf{d};\boldsymbol{\lambda}) \coloneqq \frac{1}{N} \sum_{i=1}^N \left\{ \ln q(\boldsymbol{\theta}^{(i)}|\mathbf{y}^{(i)}; \boldsymbol{\lambda}) - \ln p(\boldsymbol{\theta}^{(i)}) \right\},
     \label{e:Ud_lower_MC}   \\
    U_U(\mathbf{d};\boldsymbol{\lambda}) &\approx \widehat{U}_U(\mathbf{d};\boldsymbol{\lambda}) \coloneqq
    \frac{1}{N} \sum_{i=1}^N \left\{ \ln p(\mathbf{y}^{(i)}|\boldsymbol{\theta}^{(i)}, \mathbf{d}) - \ln q(\mathbf{y}^{(i)}; \boldsymbol{\lambda})  \right \},
\label{e:Ud_upper_MC}
\end{align}
where $\boldsymbol{\theta}^{(i)} \sim p(\boldsymbol{\theta})$ and $ \mathbf{y}^{(i)} \sim p(\mathbf{y}| \boldsymbol{\theta}^{(i)},\mathbf{d})$. In particular, 
calculating \cref{e:Ud_lower_MC} only requires \emph{sampling} the likelihood without any likelihood evaluations; it is therefore suitable to handle implicit likelihoods. 
Note that both $\widehat{U}_L$ and $\widehat{U}_U$ are not bounds themselves, but are (unbiased) \emph{estimators} of the bounds $U_L$ and $U_U$, respectively. With respect to the true EIG $U$, then, $\widehat{U}_L$ and $\widehat{U}_U$ are biased estimators whose biases are, respectively, negative and positive.

We can also form MC estimators to the  lower bound gradients in \cref{e:UL_grad_lambda,e:UL_grad_d}, yielding
\begin{align}
    \widehat{ \nabla_{\boldsymbol{\lambda}} U_L}(\mathbf{d}; \boldsymbol{\lambda}) 
    &\coloneqq \frac{1}{N_{\text{batch}}} \sum_{i=1}^{N_{\text{batch}}}  %
    \nabla_{\boldsymbol{\lambda}}
    \ln q(\boldsymbol{\theta}^{(i)}|\mathbf{y}^{(i)}; \boldsymbol{\lambda}), \\ 
    \widehat{\nabla_{\mathbf{d}} U_L}(\mathbf{d};\boldsymbol{\lambda}) &\coloneqq \frac{1}{N_{\text{batch}}} \sum_{i=1}^{N_{\text{batch}}} \ln q_{\boldsymbol{\lambda}}(\boldsymbol{\theta}^{(i)}|\mathbf{y}^{(i)}, \mathbf{d}) 
    \nabla_{\mathbf{d}} \ln p(\mathbf{y}^{(i)}|\boldsymbol{\theta}^{(i)}, \mathbf{d}).
\end{align}
Similarly, the upper bound gradient in \cref{e:UU_grad_lambda} yields 
\begin{align}
    \widehat{ \nabla_{\boldsymbol{\lambda}} U_U}(\mathbf{d}; \boldsymbol{\lambda}) 
    \coloneqq -\frac{1}{N_{\text{batch}}} \sum_{i=1}^{N_{\text{batch}}}  
    \nabla_{\boldsymbol{\lambda}}
    \ln q(\mathbf{y}^{(i)}; \boldsymbol{\lambda}).
\end{align}
In this work, we employ stochastic gradient ascent (SGA) for optimization.
In order to control the total number of forward model runs, instead of sampling $(\boldsymbol{\theta}^{(i)},\mathbf{y}^{(i)})$ from their true distributions every instance, we first generate a pool of $N_{\text{opt}}$ sample pairs at each $\mathbf{d}$, and then subdivide them into mini-batches with size $N_{\text{batch}}$ for the SGA update iterations. 

Foster \textit{et al}. \cite{Foster_19_VBOED} initially proposed parameterizing $q(\,\cdot\,;\boldsymbol{\lambda})$ using relatively simple forms of distributions such as Gaussian, Bernoulli, Beta, and simple transformations of them. 
Below we present the use of NFs to parameterize $q(\,\cdot\,;\boldsymbol{\lambda})$, with the potential to capture a richer space of distributions and in turn to achieve tighter $U_L$ and  $U_U$ bounds.

\subsection{Normalizing flows}
\label{subsection: NFs}

An NF is an invertible mapping from a target random variable $\mathbf{X}\sim p_{\mathbf{X}}(\mathbf{x})$ to a standard normal random variable $\mathbf{Z}\sim p_{\mathbf{Z}}(\mathbf{z})=\mathcal{N}(\mathbf{0},\mathbb{I})$ of the same dimension: $\mathbf{Z}=f(\mathbf{X})$ and $\mathbf{X} = g(\mathbf{Z})$ where $g\coloneqq f^{-1}$. 
In terms of transport maps, $f$ is known as the \emph{pushforward} map and $g$ is the \emph{pullback} map~\cite{Marzouk_16_Transport}. 
In practice, we approximate $f$ via a  mapping parameterized with  $\boldsymbol{\lambda}$, which produces an approximate transformation $\widetilde{\mathbf{Z}}=f(\mathbf{X};\boldsymbol{\lambda})$ and its inverse $g(\,\cdot\, ; \boldsymbol{\lambda}) \coloneqq f(\,\cdot\, ;\boldsymbol{\lambda})^{-1}$ produces $\mathbf{X} = g(\widetilde{\mathbf{Z}}; \boldsymbol{\lambda})$. If acting on the exact standard normal $\mathbf{Z}$, then $\widetilde{\mathbf{X}} = g({\mathbf{Z}}; \boldsymbol{\lambda})$ and also ${\mathbf{Z}}=f(\widetilde{\mathbf{X}};\boldsymbol{\lambda})$. 

In general, the approximate mappings used in NFs are often structured as compositions of successive simple invertible mappings: $f(\widetilde{\mathbf{X}};\boldsymbol{\lambda}) = f_n \circ f_{n-1} \circ ... \circ f_1(\widetilde{\mathbf{X}}) = f_n(f_{n-1}(...(f_1(\mathbf{{\widetilde{\mathbf{X}}}}))...))$ and $g({\mathbf{Z}};\boldsymbol{\lambda})=g_1 \circ ... g_{n-1}\circ g_n({\mathbf{Z}}) = g_1(g_{2}(...(g_n(\mathbf{{\mathbf{Z}}}))...))$ with $g_i=f_i^{-1}$ and $n\geq 1$. Note that all intermediate mappings $f_i$ and $g_i$ depend on ${\boldsymbol{\lambda}}$, but we omit their subscripts to simplify notation. 
The log density of $\widetilde{\mathbf{X}}$ can be tracked via the change-of-variable formula:
\begin{align}
    \ln q_{\widetilde{\mathbf{X}}}({\widetilde{\mathbf{X}}={\mathbf{x}};\boldsymbol{\lambda}}) = \ln p_{\mathbf{Z}}(f_n \circ f_{n-1} \circ ... \circ f_1(\widetilde{\mathbf{X}}={\mathbf{x}})) + \sum_{i=1}^n \ln \left|\text{det} \frac{\partial f_i \circ f_{i-1} \circ ... f_1(\widetilde{\mathbf{X}})}{\partial {\widetilde{\mathbf{X}}}}\right|_{\widetilde{\mathbf{X}}=\mathbf{x}},
    \label{e:composing}
\end{align}
where %
$\frac{\partial f_{i}(\widetilde{\mathbf{X}}) }{\partial \widetilde{\mathbf{X}}}$is the Jacobian of $f_{i}$.
By applying successive transformations on $\mathbf{Z}$, the PDF of the resulting variable can be highly expressive~\cite{Rezende_16_Variational, Tabak_10_Density} and effective for multi-modal, skewed, or other non-standard distribution shapes.

In the context of vOED lower bound in \cref{e:UL}, we use NF-based $q(\boldsymbol{\theta}|\mathbf{y}; \boldsymbol{\lambda})$ to approximate $p(\boldsymbol{\theta}|\mathbf{y}, \mathbf{d})$.
Among a range of choices for the architecture of invertible mappings~\cite{Papamakarios_21_NFsReview, Kobyzev_20_NFsReview}, 
we adopt the coupling layers~\cite{Dinh_16_NVP} as a special type of invertible neural network (INN)~\cite{Kruse_21_Hint, Radev_20_Bayesflow} for its efficient density evaluations and sampling in both forward ($f$) and inverse ($g$) directions. %
Several papers have shown that composing coupling layers can create flexible flows \cite{Kingma_18_Glow, Prenger_19_waveglow, Ardizzone_18_INN, Kruse_21_Hint}, and recent work by Draxler \textit{et al.} \cite{Draxler_24_Universality} shows that coupling layers form a distributional universal approximator. 
The basic form of the coupling layer that completes one full transformation starts by partitioning the standard normal random vector ${\mathbf{Z}} = [\mathbf{Z}_1, \mathbf{Z}_2]^{\top}$ into two parts of approximately equal dimension---that is, $\mathbf{Z}_1 \in \RR^{n_{{\theta}_1}}$, $\mathbf{Z}_2 \in \RR^{n_{\theta_2}}$, and $n_{\theta_1}+n_{\theta_2}=n_{\theta}$---and then composes together $n=2$ transformations that transform one part at a time. This maps $\mathbf{Z}$ to an approximate target random vector $\widetilde{\boldsymbol{\Theta}}$, i.e.,
$g(\mathbf{Z};\boldsymbol{\lambda}) = g_1 \circ g_2(\mathbf{Z}) =\widetilde{\boldsymbol{\Theta}}$, and is defined as:
\begin{align}
   g_2( \mathbf{Z} ) &= \begin{bmatrix}
\widetilde{\boldsymbol{\Theta}}_1 = [\mathbf{Z}_1 - t_2( \mathbf{Z}_2)] \odot \text{exp}[-s_2( \mathbf{Z}_2 )] \\
\mathbf{Z}_2
\end{bmatrix},  \label{eq:inn_g1} \\
g_1(g_2( \mathbf{Z} )) &= \begin{bmatrix}
\widetilde{\boldsymbol{\Theta}}_1 \\
\widetilde{\boldsymbol{\Theta}}_2 = [\mathbf{Z}_2 -t_1(\widetilde{\boldsymbol{\Theta}}_1))] \odot \text{exp}(-s_1(\widetilde{\boldsymbol{\Theta}}_1)) 
\end{bmatrix}, 
\label{eq:inn_g2}
\end{align}
where $\odot$ denotes element-wise product, and $s_1, t_1 : \mathbb{R}^{n_{\theta_1}} \rightarrow \mathbb{R}^{ n_{\theta_2}}$ and $s_2, t_2 : \mathbb{R}^{n_{\theta_2}} \rightarrow \mathbb{R}^{n_{\theta_1}}$ are arbitrary functions. The parameterizations of these functions make up $\boldsymbol{\lambda}$; for instance, if these functions are represented by neural networks, then $\boldsymbol{\lambda}$ encompasses all neural networks' weight and bias parameters. 

The inverse of $g(\,\cdot\,;\boldsymbol{\lambda})$, which is $f(\widetilde{\boldsymbol{\Theta}};\boldsymbol{\lambda}) = f_2 \circ f_1(\widetilde{\boldsymbol{\Theta}})=\mathbf{Z}$, similarly involves partitioning $\widetilde{\boldsymbol{\Theta}} = [\widetilde{\boldsymbol{\Theta}}_1, \widetilde{\boldsymbol{\Theta}}_2]^{\top}$ and can be shown to be: 
\begin{align}
    f_1(\widetilde{\boldsymbol{\Theta}}) &= \begin{bmatrix}
\widetilde{\boldsymbol{\Theta}}_1 \\
\mathbf{Z}_2 = \widetilde{\boldsymbol{\Theta}}_2 \odot \exp(s_1(\widetilde{\boldsymbol{\Theta}}_1)) + t_1(\widetilde{\boldsymbol{\Theta}}_1)
\end{bmatrix},  \label{eq:inn_f1}\\
f_2(f_1(\widetilde{\boldsymbol{\Theta}})) &= \begin{bmatrix}
\mathbf{Z}_1 = \widetilde{\boldsymbol{\Theta}}_1 \odot \exp (s_2(\mathbf{Z}_2)) + t_2(\mathbf{Z}_2) \\
\mathbf{Z}_2 %
\end{bmatrix} .
\label{eq:inn_f2}
\end{align}
The Jacobians of $f_1$ and $f_2$ are triangular matrices: 
\begin{align}
    \frac{\partial f_1(\widetilde{\boldsymbol{\Theta}})}{\partial \widetilde{\boldsymbol{\Theta}}} = \begin{bmatrix}
    \mathbb{I}_{n_{\theta_1}} & \mathbf{0} \\
    \frac{\partial \mathbf{Z}_2} %
    {\partial \widetilde{\boldsymbol{\Theta}}_1} & \text{diag}(\exp(s_1(\widetilde{\boldsymbol{\Theta}}_1)))
    \end{bmatrix}, \qquad
    \frac{\partial f_2(f_1(\widetilde{\boldsymbol{\Theta}}))}{\partial (f_1(\widetilde{\boldsymbol{\Theta}}))} = \begin{bmatrix} \text{diag}(\exp(s_2(\mathbf{Z}_2))%
    & \frac{\partial \mathbf{Z}_1}{\partial \mathbf{Z}_2} \\
    \mathbf{0}  & 
    \mathbb{I}_{n_{\theta_2}} 
    \end{bmatrix},
\end{align} 
and have respective determinants $\exp(\sum_{j=1}^{n_{\theta_2}}s_1(\widetilde{\boldsymbol{\Theta}}_1)_j)$  and $\exp(\sum_{j=1}^{n_{\theta_1}}s_2(\mathbf{Z}_2)_j)$.

Multiple such complete transformations can be composed together for greater expressiveness. For example, adopting NFs with $T=3$ sets of complete transformations entails
$f(\widetilde{\boldsymbol{\Theta}};\boldsymbol{\lambda}) = (f_2 \circ f_{1})^{T3} \circ (f_2 \circ f_{1})^{T2} \circ (f_2 \circ f_{1})^{T1}(\widetilde{\boldsymbol{\Theta}})$ and $g({\mathbf{Z}};\boldsymbol{\lambda})=(g_1 \circ g_{2})^{T3} \circ (g_1 \circ g_{2})^{T2} \circ (g_1 \circ g_{2})^{T1} ({\mathbf{Z}})$.

To incorporate the $\mathbf{y}$-dependence into the approximate posteriors $q(\boldsymbol{\theta}|\mathbf{y};\boldsymbol{\lambda})$, the $s$ and $t$ functions are designed to additionally take $\mathbf{y}$ as input, leading to a form of conditional INNs (cINNs)~\cite{Padmanabha_21_cINN}. 
When the dimension of $ \mathbf{y}$ is large, 
a summary network~\cite{Radev_20_Bayesflow} that is common to all the $s$ and $t$ functions, effectively functioning as an encoder, 
can be employed to compress $\mathbf{y}$  into a lower-dimensional summary statistic (i.e., latent variables) ${ {\mathbf{y}'}}$. The parameters of the summary network then become part of $\boldsymbol{\lambda}$. The summary network can be particularly beneficial to contain the growth of parameters when $s$ and $t$ functions are represented by neural networks.
\Cref{fig: diagram} illustrates the overall cINN transformation for vOED lower bound.

\begin{figure}[htb]
   \centering
    \includegraphics[width=\textwidth]{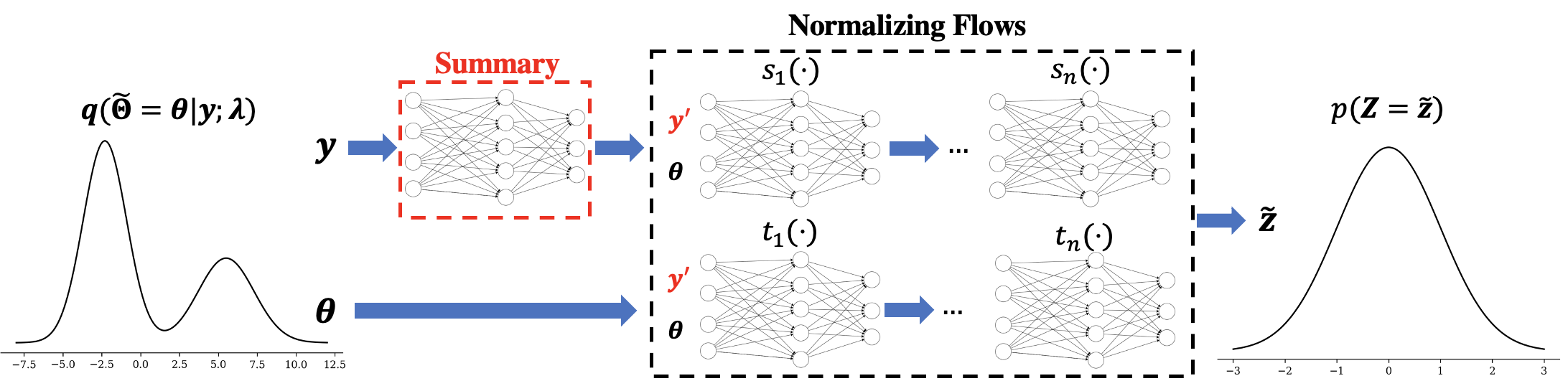}%
    \caption{Diagram for the cINN NFs transformation.}
    \label{fig: diagram}
\end{figure}

In the case of vOED upper bound in \cref{e:UU}, NF-based $q(\mathbf{y};\boldsymbol{\lambda})$ can be similarly used to approximate $p(\mathbf{y})$.
The transformation $g(\mathbf{Z};\boldsymbol{\lambda})$ then mirrors \cref{eq:inn_g1,eq:inn_g2}, while $f(\mathbf{Y};\boldsymbol{\lambda})$ mirrors  \cref{eq:inn_f1,eq:inn_f2}.

In summary, the final approach to determine the optimal design entails maximizing the EIG via its lower bound:
\begin{align}
\mathbf{d}^{\ast}, \boldsymbol{\lambda}^{\ast} = \argmax_{\mathbf{d} \in \mathcal{D}, \boldsymbol{\lambda}} \widehat{U}_L = \argmax_{\mathbf{d}, \boldsymbol{\lambda}} \frac{1}{N}   \sum_{i=1}^N \left\{\ln q(\boldsymbol{\theta}^{(i)}| \mathbf{y}^{(i)};\boldsymbol{\lambda}) - \ln p(\boldsymbol{\theta}^{(i)} )\right\},
\label{e:final_lower}
\end{align}
where $\boldsymbol{\theta}^{(i)} \sim p(\boldsymbol{\theta})$, and $\mathbf{y}^{(i)} \sim p(\mathbf{y}^{(i)}|\boldsymbol{\theta}^{(i)}, \mathbf{d})$, and $\boldsymbol{\lambda}$ represents all the neural network weight and bias parameters in the cINN and summary network.
For results that show upper bound estimates, they are based on solving the problem:
\begin{align}
    \mathbf{d}^{\ast} = \argmax_{\mathbf{d} \in \mathcal{D}} \min_{\boldsymbol{\lambda}} \widehat{U}_U = \argmax_{\mathbf{d} \in \mathcal{D}} \min_{\boldsymbol{\lambda}} \frac{1}{N}   \sum_{i=1}^N \left\{\ln p(\mathbf{y}^{(i)}|\boldsymbol{\theta}^{(i)}, \mathbf{d}) - \ln q(\mathbf{y}^{(i)}; \boldsymbol{\lambda})\right\}.
    \label{e:final_upper}
\end{align}

\section{Numerical Experiments}
\label{section: Applications}
We demonstrate vOED-NFs across four design cases involving models of varying complexity: (1) a nonlinear model with 3 parameters and 1 design variable; (2) a linear model with 21 parameters and 400-dimensional design vector; (3) a PDE model for cathodic electrophoretic deposition; and (4) a stochastic model of aphid population with an implicit likelihood. 

\subsection{Case 1: Low-dimensional nonlinear design}
\label{subsection: case1}
Consider a nonlinear observation model\footnote{There is a typo to (32) in the journal version of this paper. The equation provided here is the correct one and consistent with our results and code.}
\begin{align}
    y = G(\boldsymbol{\theta}, d) + \epsilon \qquad \text{where} \qquad
    G(\boldsymbol{\theta}, d) = \theta_1^3 d^2 + \theta_2 e^{-|0.2-d|} + \sqrt{\theta_3^2 d \times 2},
\end{align}
with independent prior on parameters  $\theta_1 \sim \CN(0.5, 0.3^2), \theta_2 \sim \CN(0.3, 0.7^2), \theta_3 \sim \CN(0.5, 0.8^2)$; $d \in [0, 1]$; and measurement noise $\epsilon$ following a Gaussian mixture distribution:
\begin{equation}
     p(\epsilon) = 
     \frac{1}{2\sqrt{2 \pi \sigma_0^2}} \exp\left(-\frac{(\epsilon - \mu_1)^2}{2\sigma_0^2}\right) + \frac{1}{2 \sqrt{2 \pi \sigma_0^2}} \exp \left(-\frac{(\epsilon - \mu_2)^2}{2\sigma_0^2}\right),
     \label{e: epsilon}
\end{equation}
where $\mu_1 = 0.1, \mu_2 = -0.1, \sigma = 0.05$.

\Cref{fig:Toy1_a} presents the EIG estimates across the discretized design space using (a) (Hi-NMC) a high-quality $\widehat{U}_{\text{NMC}}$ with $N_{\text{out}}=2 \times 10^4 $ and $N_{\text{in}}=2 \times 10^4 $ samples, equating to $4 \times 10^8$ forward model evaluations total, (b) (Lo-NMC) a low-quality $\widehat{U}_{\text{NMC}}$ with $N_{\text{out}}=200$ and $N_{\text{in}}=200$ samples, equating to $4 \times 10^4$ forward model evaluations total, (c) (Re-NMC) a NMC estimator with  $N_{\text{out}}=N_{\text{in}}=2 \times 10^4 $, where the inner loop reuses the outer loop samples, equating to $2 \times 10^4 $ forward model evaluations total, (d) (vOED-G) EIG lower bound $\widehat{U}_L$ using Gaussian distributions optimized with $N_{\text{opt}}=2 \times 10^4 $ samples, and (e) (vOED-NFs) EIG lower bound $\widehat{U}_L$ using NFs optimized with $N_{\text{opt}}=2 \times 10^4 $ samples. Furthermore, the lower bound estimate $\widehat{U}_L$ at each $d$ is evaluated using $N=1 \times 10^4 $ samples. For (d) and (e), the total number of forward model evaluation is therefore $2 \times 10^4 + 1 \times 10^4  = 3 \times 10^4 $.
vOED-NFs adopts $T=5$ sets of complete transformations of the cINN  described in \cref{subsection: NFs}.
Additional training details and hyperparameter choices for (d) and (e)
can be found in \ref{subsection: case1_hyperparam}.

In \cref{fig:Toy1_a}, Hi-NMC is designated as the reference solution. In comparison to this reference, Lo-NMC exhibits moderate bias while vOED-NFs performs extremely well. Re-NMC yields near-identical results as Hi-NMC; while faster, it still requires an explicit likelihood and does not provide approximate posterior distributions, just like the other NMC methods. vOED-G, due to its much simplistic Gaussian variational distributions, deviates significantly from the expected utility trend and misidentifies the optimal design. 
However, the training cost for vOED-NFs is also higher than vOED-G, as illustrated through the training convergence history for $d=1.0$ in \cref{fig:Toy1_b}. Here vOED-G shows a very rapid convergence to $\boldsymbol{\lambda}^*$, while vOED-NFs takes longer to stabilize and experiences greater fluctuation. 
This timing difference induced by optimization is negligible when the overall computations are dominated by the forward model runs, but can become significant when the forward model is inexpensive. Additional information regarding the time required for EIG estimation is given in \ref{section: training_time}.

\begin{figure}[htb]
 \centering
  \subfloat[EIG estimates with different estimators]{\includegraphics[width=0.43\textwidth]{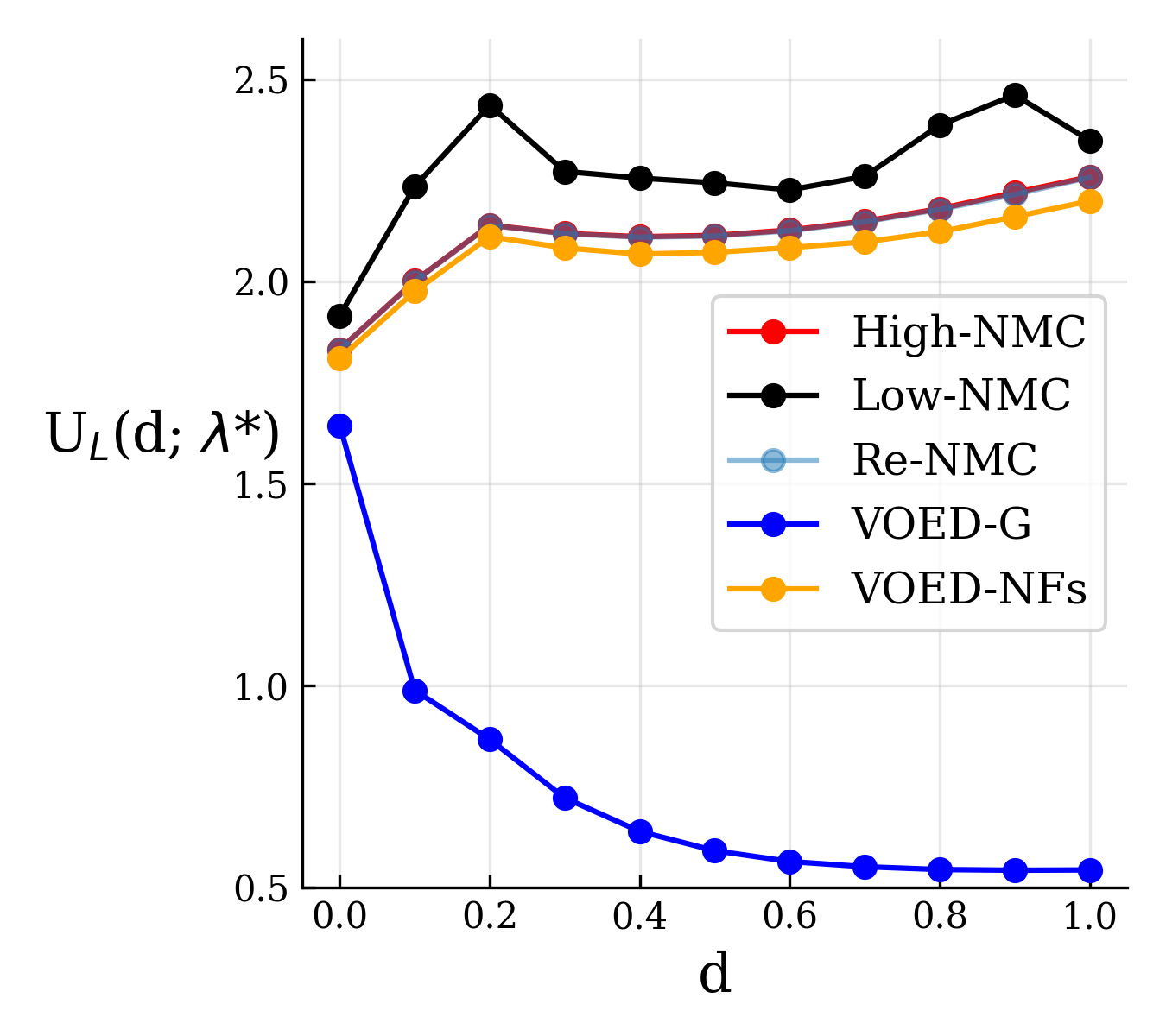}
  \label{fig:Toy1_a}}
  \subfloat[Training convergence at $d = 1$]{\includegraphics[width=0.48\textwidth]{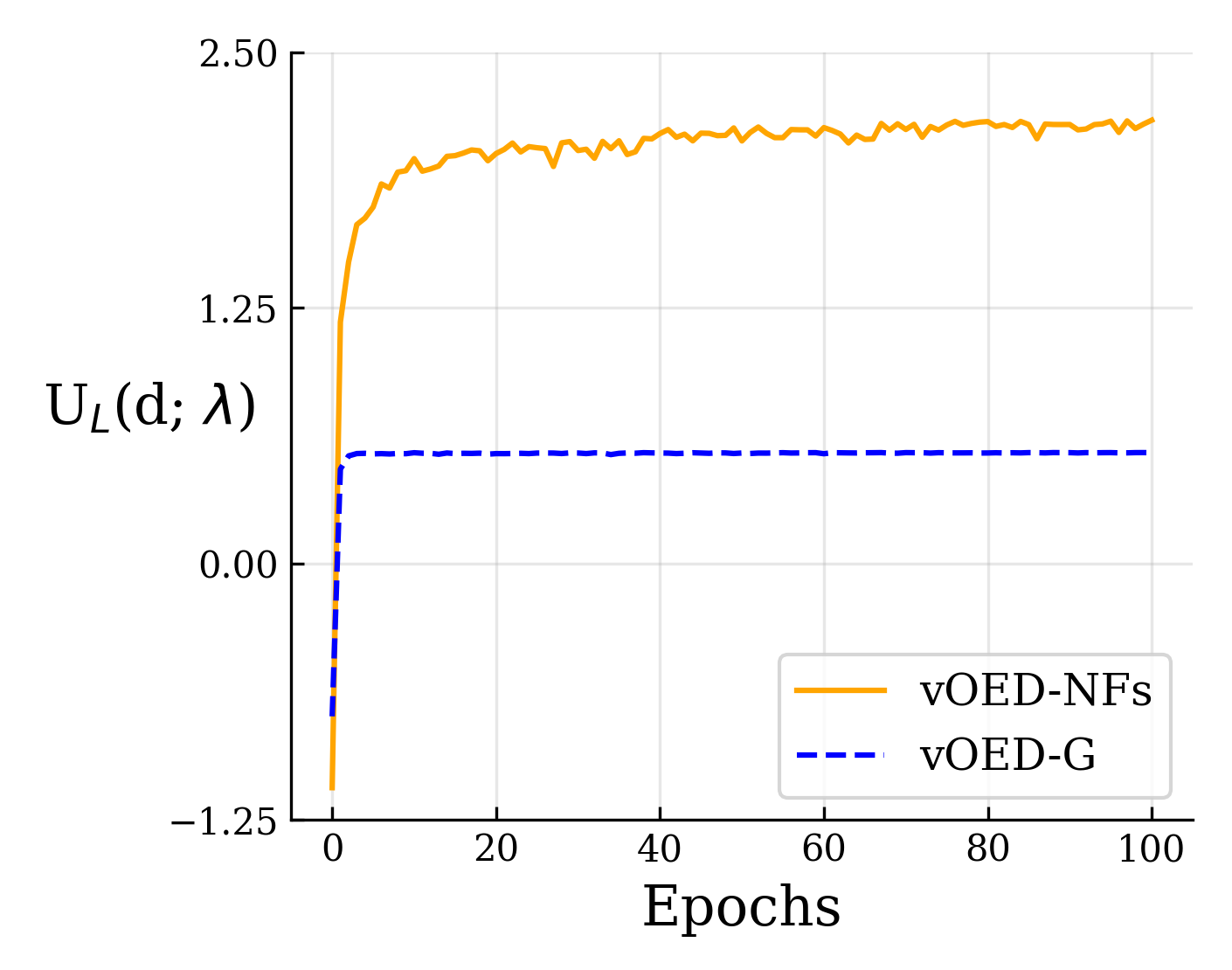}\label{fig:Toy1_b}}
  \caption{Case 1. EIG estimates and sample training convergence plot. The Re-NMC plot line is right on top of the High-NMC plot line, as they produce near-identical results.} 
  \label{fig: Toy 1}
\end{figure}

To further analyze the ability of vOED-NFs in approximating the true posteriors, we compare the resulting variational posteriors to a high-quality reference posterior obtained using sequential Monte Carlo (SMC)~\cite{Doucet_13_SMC} at two designs: $d=0.2$ in \cref{fig: Comparison of Posteriors at d = 0.2} and $d=1.0$ in \cref{fig: Comparison of Posteriors at d = 1}. Each column plots the marginal posteriors of the three parameters $q_{\boldsymbol{\lambda}^{\ast}}(\theta_1|y, d)$,
$q_{\boldsymbol{\lambda}^{\ast}}(\theta_2|y, d)$,
$q_{\boldsymbol{\lambda}^{\ast}}(\theta_3|y, d)$,
while each row corresponds to a different sample of observed $y$.
The vOED-NFs posteriors approximate the SMC reference posteriors very well, even with multi-modal posteriors as seen in \cref{fig: Comparison of Posteriors at d = 1} for $\theta_3$.

\begin{figure}[b!]
 \centering
 \includegraphics[width=\textwidth]{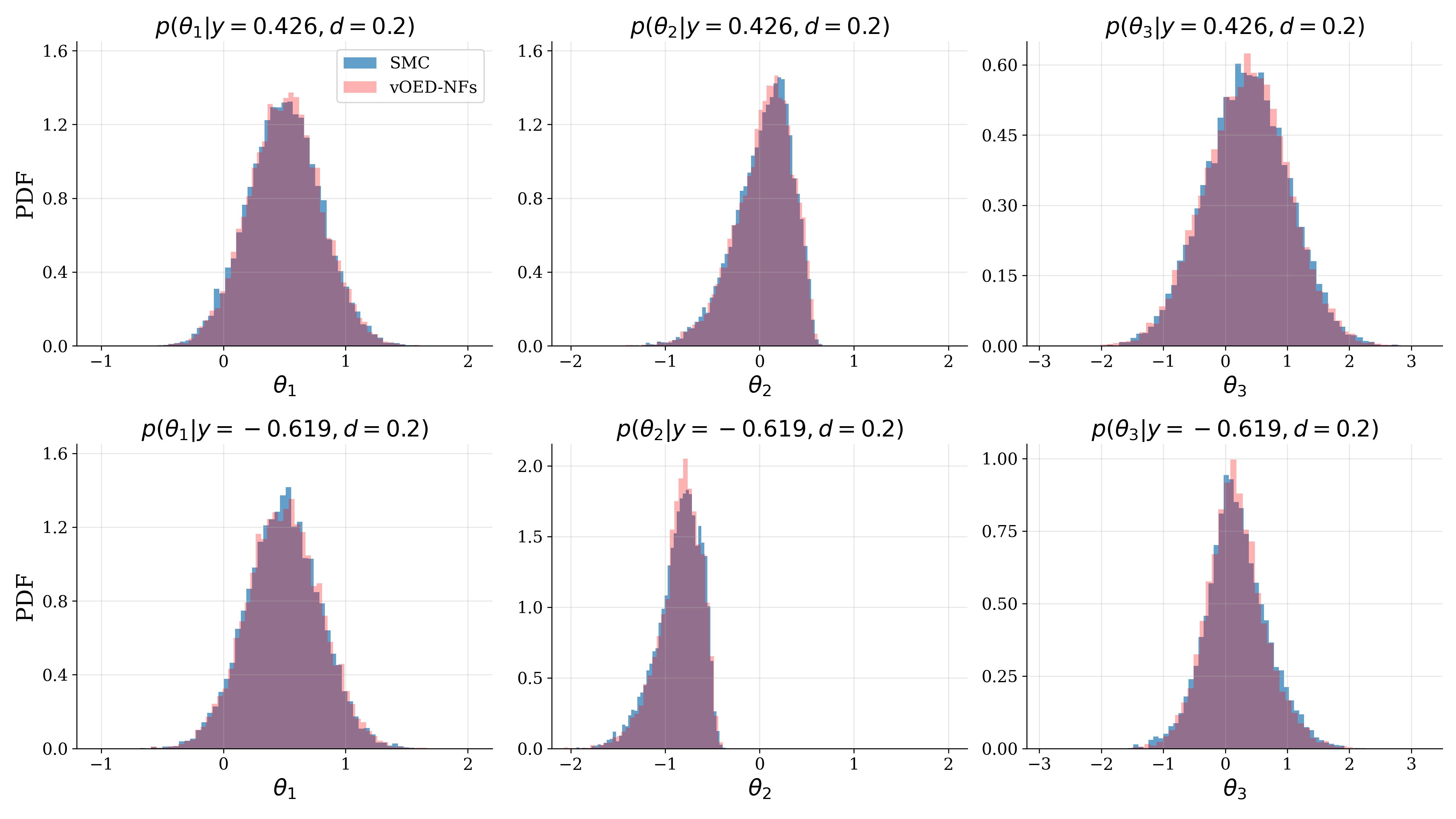}
   \caption{Case 1. Comparison of marginal posteriors obtained from SMC and vOED-NFs  at $d = 0.2$.}
  \label{fig: Comparison of Posteriors at d = 0.2}
\end{figure}

\begin{figure}[htb]
  \centering
  \includegraphics[width=\textwidth]{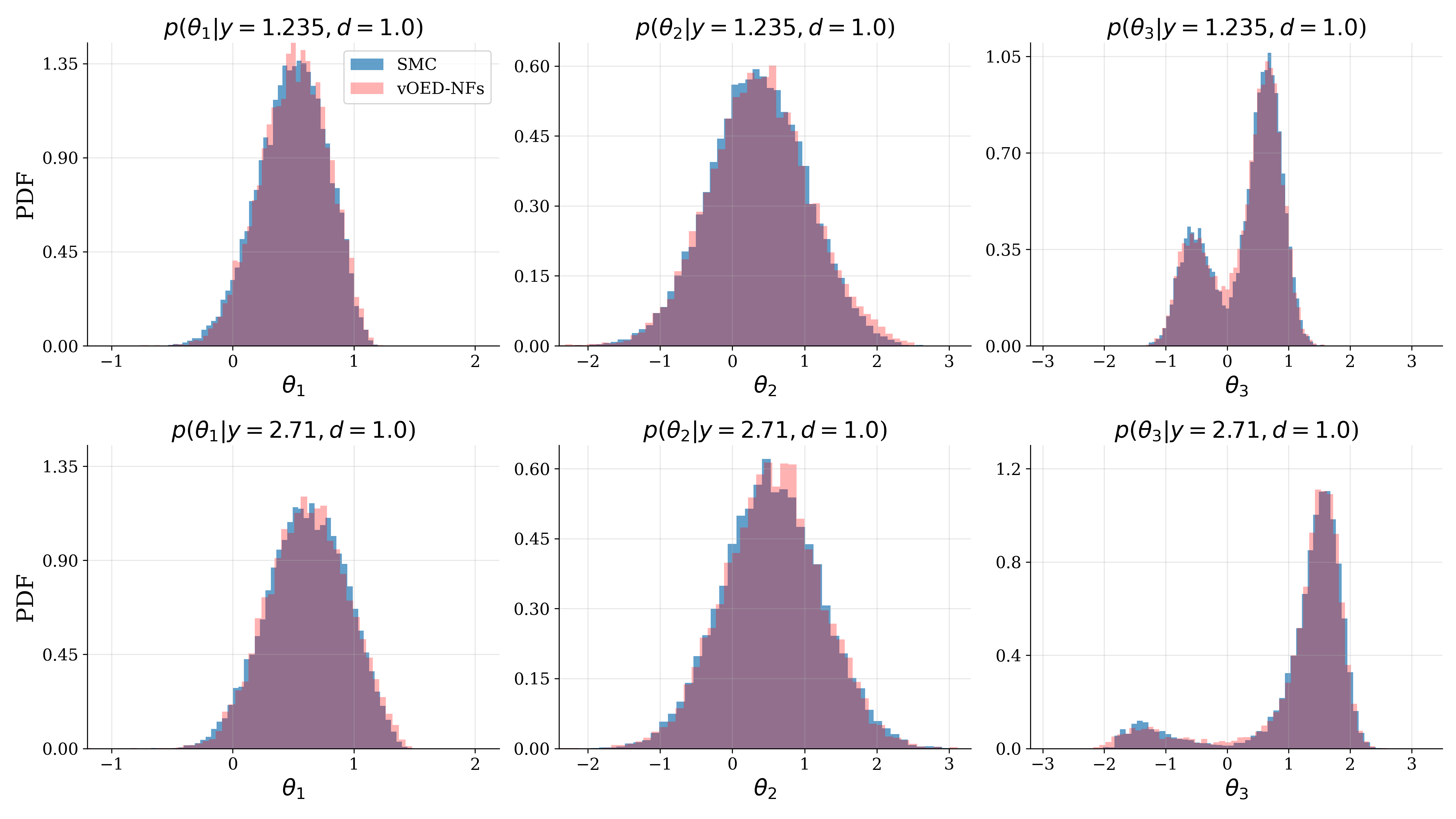}
  \caption{Case 1. Comparison of marginal posteriors obtained from SMC and vOED-NFs at $d = 1.0$.}
  \label{fig: Comparison of Posteriors at d = 1}   
\end{figure}

\Cref{fig: Assess cost and flexibility of different transformation} investigates the impact of the number of transformations and optimization sample size $N_{\text{opt}}$ in vOED-NFs. The plots show the EIG lower bound estimates $\widehat{U}_L$ when adopting $T=\{1, 3, 5, 7\}$ sets of complete cINN transformations, and using $N_{\text{opt}}=\{5 \times 10^3,  10^4, 5 \times 10^4\}$. The EIG lower bound estimates are still evaluated with $N= 10^4$ samples. In general, employing 3 transformations is adequate to identify the optimal design $d^{\ast}=1$. 
However, when $N_{\text{opt}}$ is small, adding more transformations does not provide significant improvement. When $N_{\text{opt}}$ is increased, the value from adding transformations is more notable and stabilized around $T=5$; this is consistent with previous work \cite{Radev_20_Bayesflow} that reported composing more than 5 transformations provides only slight additional benefits. 
This diminishing return is likely due to the much larger number of NFs parameters when a high number of transformations is used, making them more prone to overfitting.
Additional testing and discussions for the $N_{\text{opt}}=5 \times 10^3$ case can be found in \ref{subsection: case1_hyperparam}.

\begin{figure}[htb]
 \centering
  \includegraphics[width=\textwidth]%
  {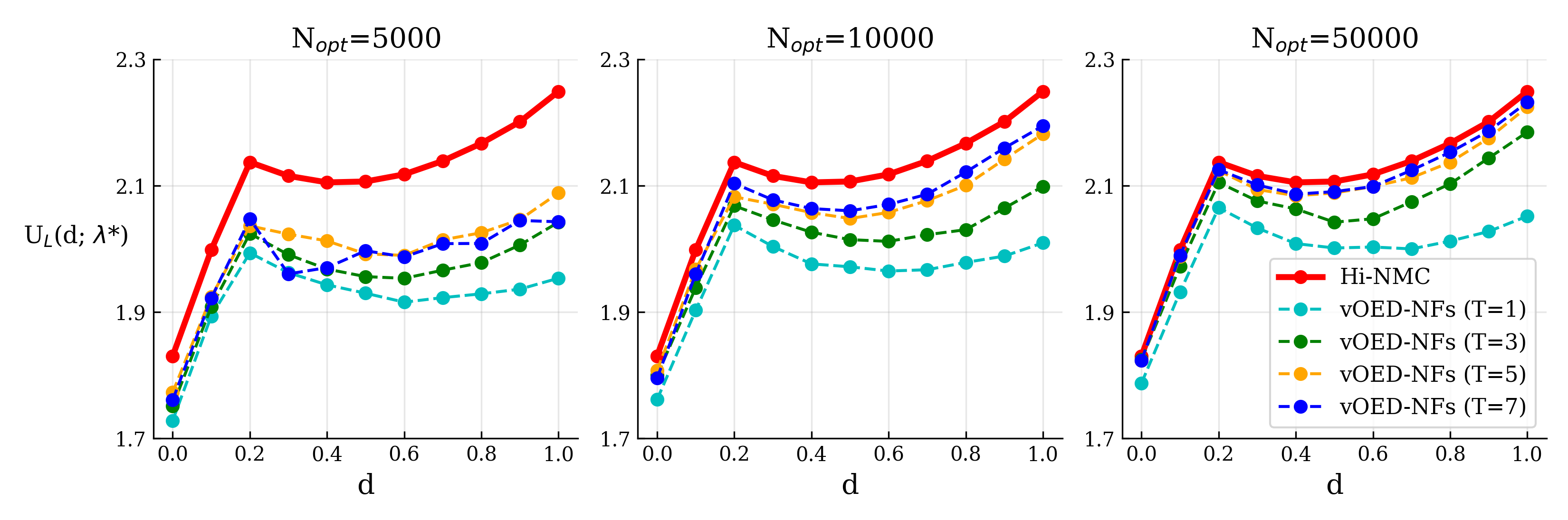}
  \caption{Case 1. EIG under different number of transformations $T$ and optimization sample sizes $N_{\text{opt}}$.}
  \label{fig: Assess cost and flexibility of different transformation}
\end{figure}

\subsection{Case 2: High-dimensional linear design}
\label{subsection: case2}

This next example showcases the joint optimization of $\mathbf{d}$ and $\boldsymbol{\lambda}$ and in a higher dimensional setting. The setup follows the example in~\cite{Foster_20_Unified}, with an observation model:
\begin{align}
    y_j  = \mathbf{d}_j^{\top}  \mathbf{w} + \epsilon_j,  \qquad j = 1, \ldots, n,
\end{align}
where $y_j\in \mathbb{R}$ is the $j$th observation, $\mathbf{d}_j\in \mathbb{R}^{p}$ is its corresponding design vector, $\epsilon_j \sim \CN(0, \sigma^2)$ is a Gaussian noise random variable, and
$\boldsymbol{\theta} = \{\mathbf{w}, \sigma\}$ encompasses the $p$-dimensional regression coefficient vector $\mathbf{w}$ and noise standard deviation $\sigma$. We set $p=20$ and $n=20$, hence the parameter dimension is 20 and the total design dimension is 400. 
Independent priors are adopted for $\mathbf{w}_j \sim \text{Laplace}(0,1)$ and $\sigma \sim \text{Exp}(1)$. 
Lastly, a design constraint $||\mathbf{d}_j||_1=1$ is imposed to reflect a resource budget; this is implemented through re-normalization of $\mathbf{d}$ after each SGA update in order to map it back to the feasible region.
During the joint optimization of $\mathbf{d}$ and $\boldsymbol{\lambda}$, the total number of forward model runs (i.e., the total number of $(\boldsymbol{\theta}^{(i)}$, $\mathbf{y}^{(i)})$ pairs across all design epochs) is fixed to $10^6$. This is the same computational budget used by~\cite{Foster_20_Unified} that adopted Gaussian and Gamma variational distributions for $\mathbf{w}$ and $\sigma$, respectively; we refer to the setup in~\cite{Foster_20_Unified} as vOED-GG. 
Further details on the computational setup can be found in~\ref{subsection: case2_hyperparam}.

Optimization results are presented in \cref{table: regression results}, with the middle column showing the mean $\pm$ one standard deviation of 10 lower bound estimates $\widehat{U}_L$ (each evaluated with $N = 10^4$ samples) at the optimal design found by vOED-GG and vOED-NFs, and the right column showing the mean $\pm$ one standard deviation of 10 high-quality NMC estimates (each using $N_{\text{out}} = 10^4$ and $N_{\text{in}}  = 10^4$) at each method's optimal design to provide an accurate EIG estimate and using a common estimation method. We observe that vOED-NFs's optimal design achieves a tighter lower bound estimate and also a higher NMC estimate, indicating that a better design has been found than vOED-GG. However, the vOED-NFs lower bound estimate is not tight, and the remaining discrepancy between $\widehat{U}_L$ and $\widehat{U}_{\text{NMC}}$ may be contributed by a number of factors: (1) the functional form of cINN; (2) the number of cINN transformations; (3) the functional representation of $s$'s and $t$'s; (4) the possibility to converge to a local optimum during optimization; and (5) the finite sample sizes.

\begin{table}[htb]
\centering
 \begin{tabular}{|c |c | c|} 
 \hline
 Method & $\widehat{U}_L(\mathbf{d}^*;\boldsymbol{\lambda^*})$ & $\widehat{U}_{\text{NMC}}(\mathbf{d}^*)$ \\
\hline
\hline
vOED-GG \; & \; 12.07  (0.11) \; & \; 24.28 (0.24) \; \\
\hline
vOED-NFs  \;  & \; 20.94 (0.19) \; & \; 24.91 (0.24) \; \\
\hline
\end{tabular}
\caption{Case 2. Middle column shows the mean of 10 lower bound estimates $\widehat{U}_L$ (each is evaluated with $N = 10000$ samples) at the optimal design found by vOED-GG and vOED-NFs, and the right column shows the mean of 10 high-quality NMC estimates (each with $N_{\text{out}} = 10000$ and $N_{\text{in}}  = 10000$) at the optimal designs. The $(\cdot)$ values are one standard deviation estimates for the estimators.}
\label{table: regression results}
\end{table}

In order to compare the posterior approximation capabilities of the two approaches, we look at their approximate posteriors at a common design with each $\mathbf{d}_j=\mathbf{e}_j$ set to the standard unit vector in the $j$th dimension, and at three $\mathbf{y}$ samples. \Cref{fig: regression posterior-AF,fig: regression posterior-NF} present the marginal posteriors for $\{w_1, w_2, w_3, w_4, \sigma\}$ using No-U-Turn Hamiltonian Monte Carlo (NUTS-HMC)~\cite{Hoffman_14_NoUTurn} as the reference posterior distribution, along with the approximations obtained from vOED-GG and vOED-NFs. From the figure, it is evident that the vOED-NFs posteriors can capture the non-Gaussian structures much better than vOED-GG's Gaussian/Gamma distributions, which is confirmed by the empirical KL divergence from the NFs posteriors to MCMC posteriors. 

\begin{figure}[htb]
    \centering
    \includegraphics[width=\textwidth]
    {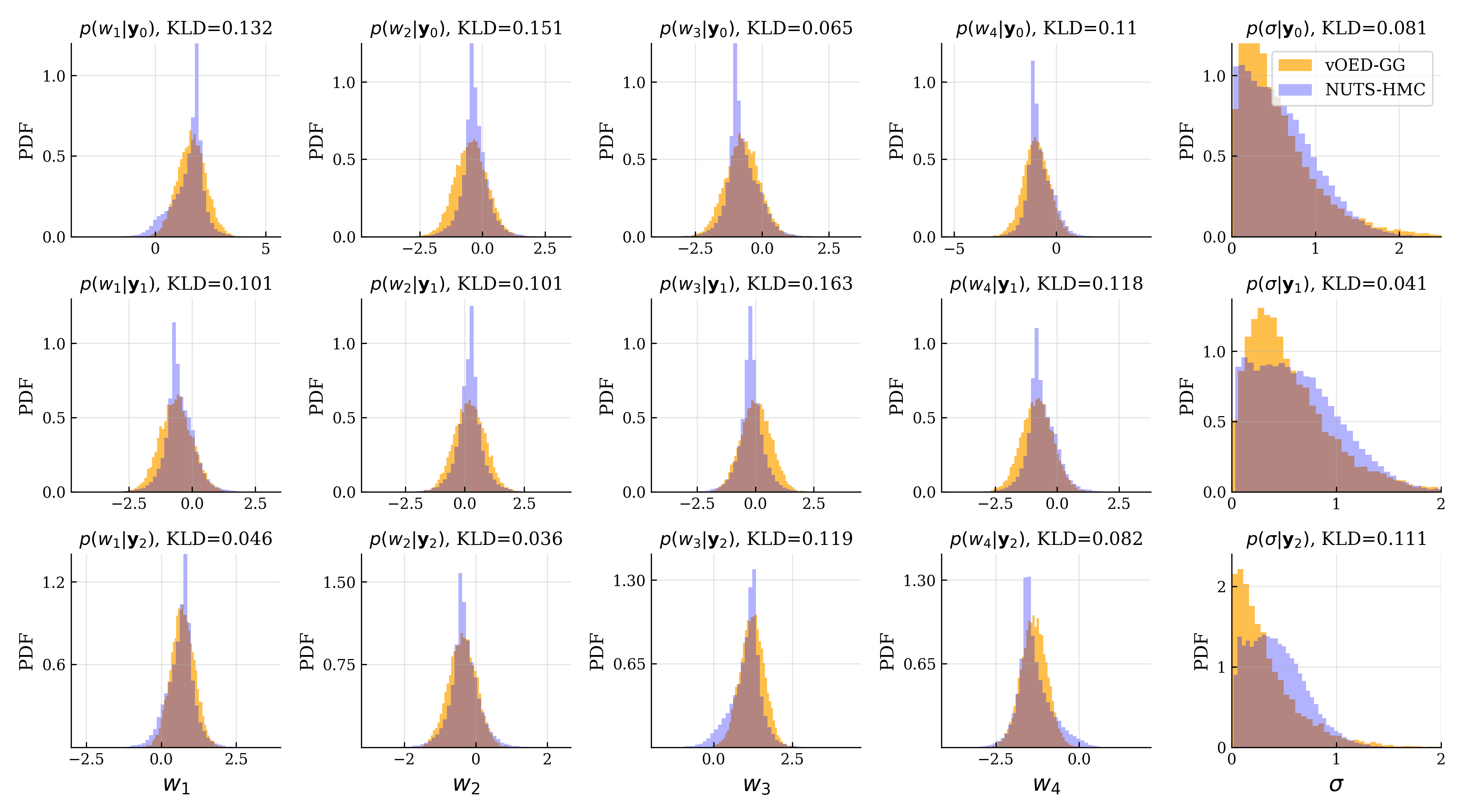}
    \caption{Case 2. Comparison of marginal posteriors obtained from NUTS-HMC and vOED-GG~\cite{Foster_19_VBOED} at design where $\mathbf{d}_j=\mathbf{e}_j$, with KLD denotes the empirical KL divergence from NFs posteriors to MCMC posteriors.}
    \label{fig: regression posterior-AF}
\end{figure} 

\begin{figure}[h!]
    \centering
    \includegraphics[width=\textwidth]
    {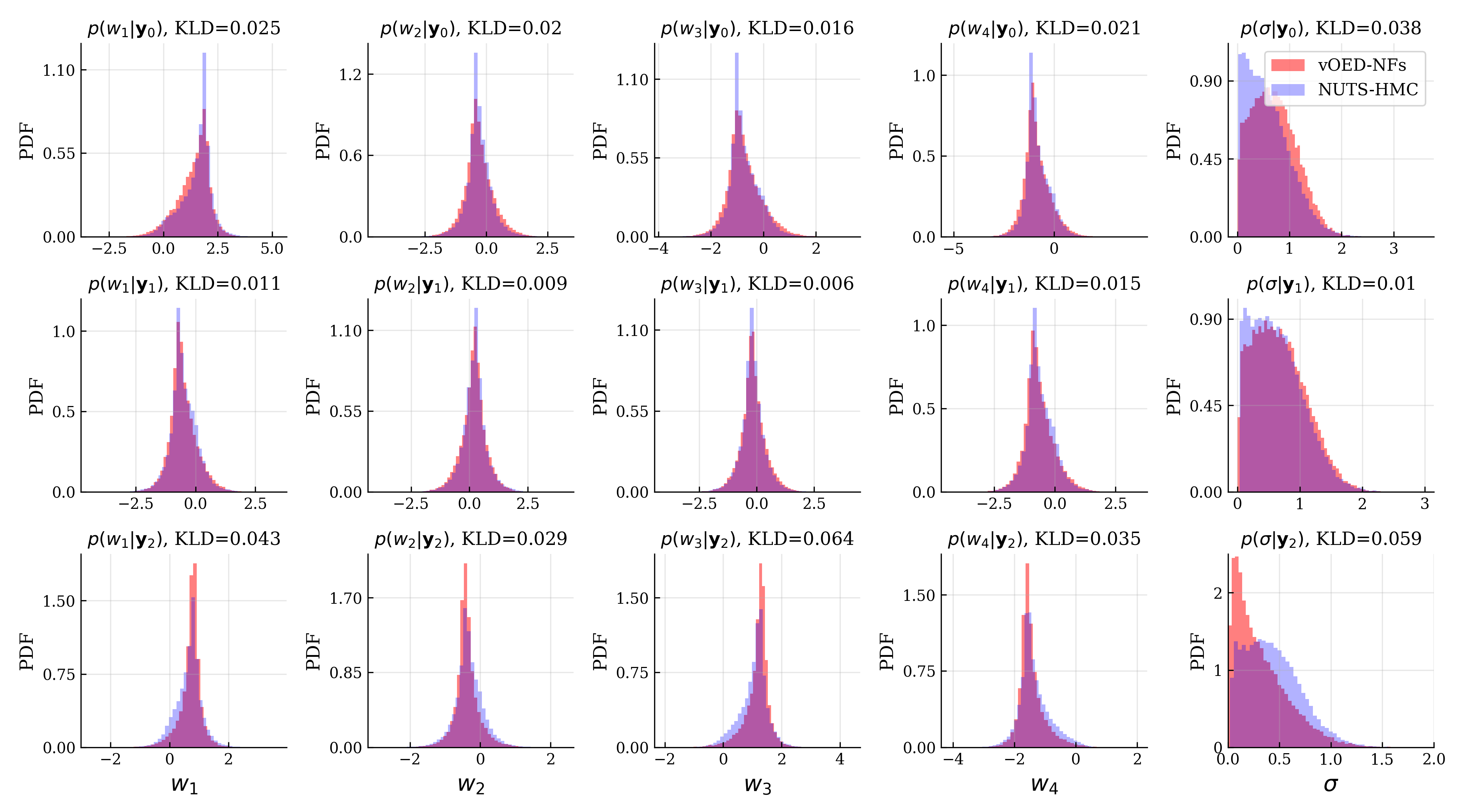}
    \caption{Case 2. Comparison of marginal posteriors obtained from NUTS-HMC and vOED-NFs~\cite{Foster_19_VBOED} at design where $\mathbf{d}_j=\mathbf{e}_j$, with KLD denotes the empirical KL divergence from NFs posteriors to MCMC posteriors.}
    \label{fig: regression posterior-NF}
\end{figure} 

\subsection{Case 3: Cathodic electrophoretic deposition (e-coating)}
\label{subsection: case3}

Cathodic electrophoretic deposition, commonly known as e-coating, is a technique for applying protective coatings to automobile surfaces. 
During the procedure, an anode and a cathode (the car body) are placed in a colloidal bath in which the electric current triggers an electrochemical reaction resulting in the deposition of material on the car body. A proper coating of the car body is achieved when a sufficient thickness of coating is reached. Here we apply vOED-NFs to design e-coating experiments, and employ both the lower and upper bound setups respectively in \cref{e:UL_bound,e:UU_bound}.

The physical model adopted for the e-coating process, consisting of a PDE, is the baseline model from~\cite{Jacobsen_24_Enhancing} and also summarized in \ref{subsection: case3_Forward_Model}. The parameters of interest, $\boldsymbol{\theta} = \{ C_v, j_{\min}, Q_{\min} \}$, are the volumetric coulombic efficiency (m$^3$/C), minimum current (A/m$^2$), and minimum charge (C/m$^2$), respectively. Each parameter is endowed with a truncated normal prior, with detailed prior parameters provided in \ref{subsection: case3_Forward_Model}. A measurement on the current (mA) can be obtained at time $t$ (s), modeled as
\begin{align}
 y_t = j(\boldsymbol{\theta}, j_0, t)(1 + \epsilon),
\end{align}
where $j(\boldsymbol{\theta}, j_0, t)$ is the forward model mapping from parameters to observables at time $t$ and under constant current boundary condition $j_0$ (mA), and $\epsilon \sim \mathcal{N}(0,  0.1^2)$ represents a relative measurement noise.
The design variable is $j_0$, and three candidate designs are considered: 
(1) $j_0=10$ mA with measurements made at $t=\{10, 20,\ldots,100\}$ s; 
(2) $j_0=7.5$ mA with measurements made at $t=\{20, 40,\ldots,200\}$ s; and
(3) $j_0=5.0$ mA with measurements made at $t=\{30, 60,\ldots,300\}$ s.
Hence, all designs always involve $n_{y}=10$ measurements at regular intervals; the lower $j_0$ designs have slower decays of the current (e.g., see Figure D.14 in~\cite{Jacobsen_24_Enhancing}) and thus longer intervals are adopted.

\Cref{fig: 10-dim result} presents the convergence of the upper and lower bound estimates from \cref{e:Ud_lower_MC,e:Ud_upper_MC}, along with a reference EIG value,
when optimizing over $\boldsymbol{\lambda}$ at design $j_0=5.0$ mA.
The reference EIG is obtained from a high-quality NMC with the `reuse' technique described in~\cite{Huan_13_Simulation} in order to accommodate the relatively high computational cost for each forward model evaluation (i.e., a PDE solve). Thus, even with $N_\text{out} = N_\text{in} = 10^5$, the reuse technique only requires $10^5$ total forward model evaluations (instead of $10^{10}$, which would be impractical). 
We further provide comparison of the bounds between low-quality optimization that uses $N_{\text{opt}} = 10^4$ versus high-quality estimation that uses $N_{\text{opt}} = 8\times10^4$ (the evaluation of the bounds are always performed with  $N = 10^4$). 
Results indicate that increasing $N_{\text{opt}}$ tightens both bounds. Under the same $N_{\text{opt}}$, the upper bound appears to be tighter than the lower bound. 
To illustrate the ability of $q(\mathbf{y};\boldsymbol{\lambda})$ in representing the marginal likelihood $p(\mathbf{y}|\mathbf{d})$, their marginal distributions are shown in \cref{fig: 10-dim result distribution}, where we see excellent agreement between the approximation and the true distributions obtained from direct Monte Carlo.  %

\begin{figure}[htb]
   \centering\includegraphics[width=0.65\textwidth]
    {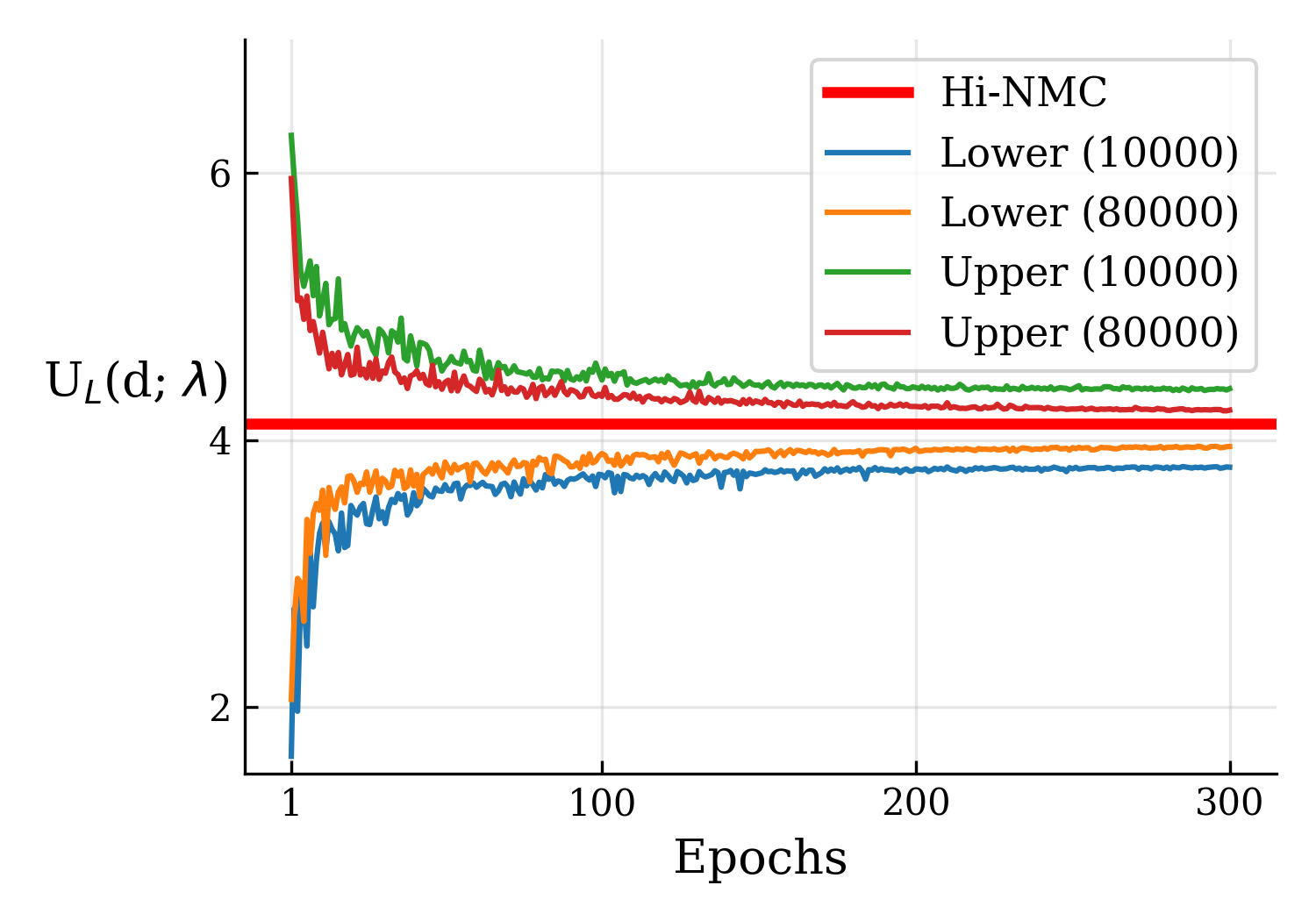}
    \caption{Case 3. Convergence history of the lower and upper bounds at $d=j_0 = 5$ mA and 10 observations when optimizing $\boldsymbol{\lambda}$ using different $N_{\text{opt}}$. Hi-NMC is the high-quality reference EIG estimate. Here the upper bound estimates appear to be tighter than the lower bound estimates. }
    \label{fig: 10-dim result}
\end{figure}

\begin{figure}[htb]
   \centering
    \includegraphics[width=\textwidth]
    {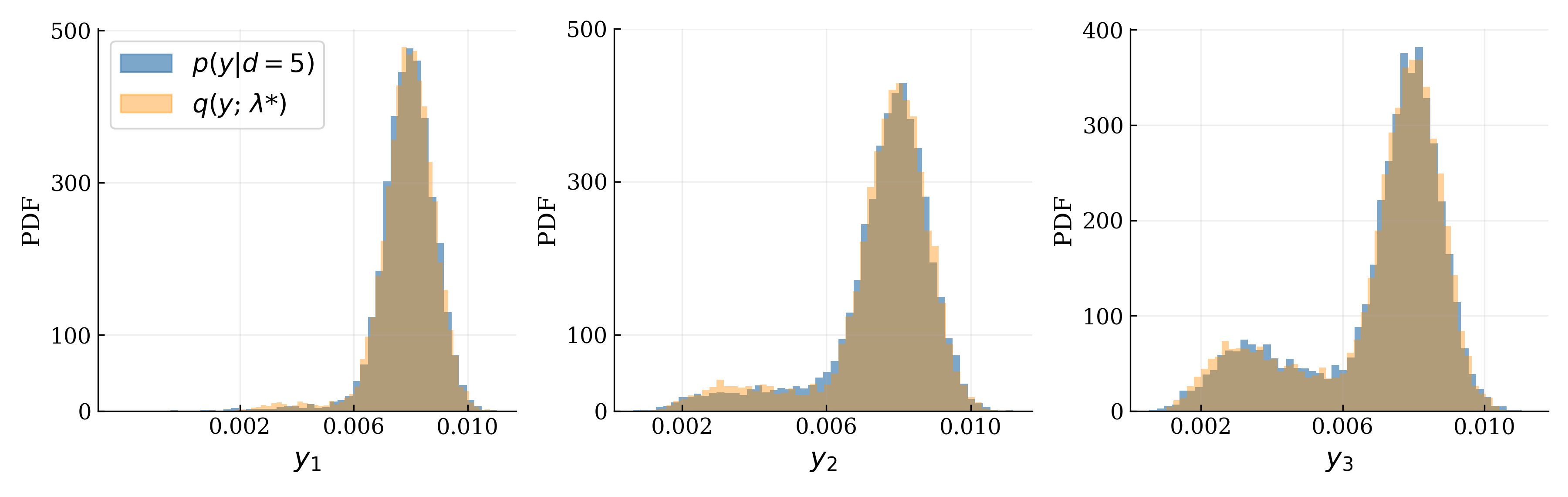}
    \caption{Case 3. Marginal distributions $p(y_1|d), p(y_2|d), p(y_3|d)$ at $d=j_0=5$ mA obtained using MC sampling and vOED-NFs with $N_{\text{opt}} = 10000$.}
    \label{fig: 10-dim result distribution}
\end{figure} 

In scenarios where the dimension of $\boldsymbol{y}$ is high, the input size to the $s$ and $t$ networks of the cINN becomes large. To combat this high dimensionality, summary networks are employed to convert $\mathbf{y}$ to a lower-dimensional $\mathbf{y}'$, specifically through a 
long short-term memory (LSTM) network~\cite{LSTM2000} as recommended in~\cite{Radev_20_Bayesflow} to accommodate the sequential nature of $\mathbf{y}$. 
\Cref{fig: Ecoat_lower_bounds_convergence} presents the same convergence of bound estimates but now with $n_y=50$ measurements taken in each experiment, i.e., $\boldsymbol{y} = \{y_{t_1}, \ldots, y_{t_{50}}\}$, and with a summary network that compresses it to a 10-dimensional $\mathbf{y}'$. The figure indicates that the lower bound estimates are now tighter than their upper bound counterparts. 
Together with \cref{fig: 10-dim result}, these observations suggest that when the dimension of $\mathbf{y}$ significantly exceeds that of $\boldsymbol{\theta}$, EIG may be more accurately estimated though the lower bound estimate that approximates the posterior, than the upper bound estimate that approximates the marginal likelihood. 

Various lower bound estimates, using vOED-G, vOED-NFs, vOED-NFs-LSTM, and the high-quality NMC, are computed for all three candidate designs and shown in~\cref{fig: Ecoat_lower_bounds}. All results indicate $j_0 = 0.75$ mA achieving the best design, although vOED-G incurs a notable error in estimating the EIG compared to the rest. The vOED-NFs lower bound estimate is further tightened when the LSTM summary network is used. The vOED-NFs-LSTM using different $N_{\text{opt}}$ also further supports the benefit when optimizing using a larger sample size. 

\begin{figure}[htp]
   \centering
    \includegraphics[width=0.65\textwidth]{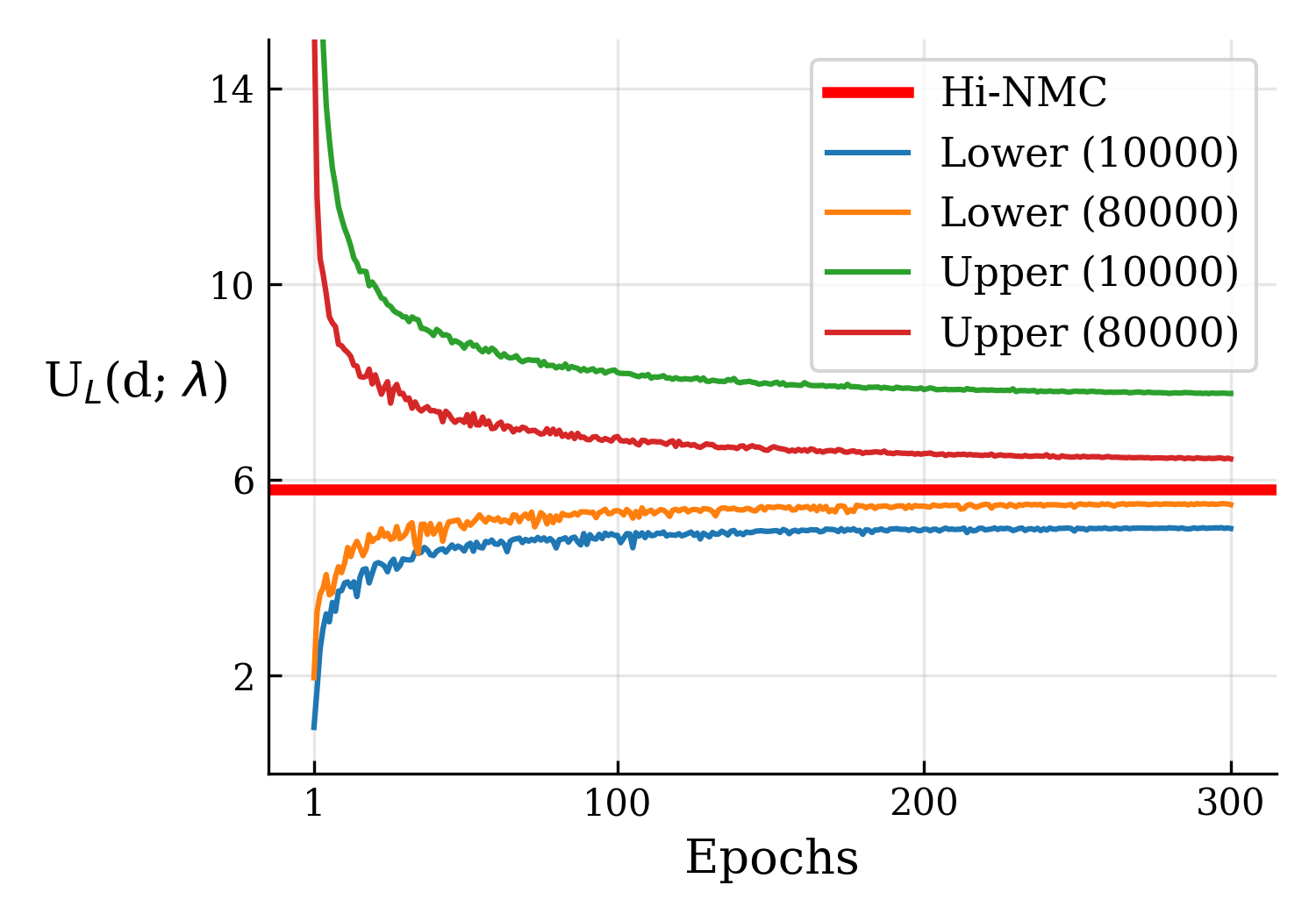}
    \caption{Case 3. Convergence history of the lower and upper bounds at $d=j_0 = 5$ mA and 50 observations when optimizing $\boldsymbol{\lambda}$ using different $N_{\text{opt}}$. Hi-NMC is the high-quality reference EIG estimate. Here the lower bound estimates appear to be tighter than the upper bound estimates. } 
    \label{fig: Ecoat_lower_bounds_convergence}
\end{figure} 

\begin{figure}[htp]
   \centering\includegraphics[width=0.65\textwidth]
    {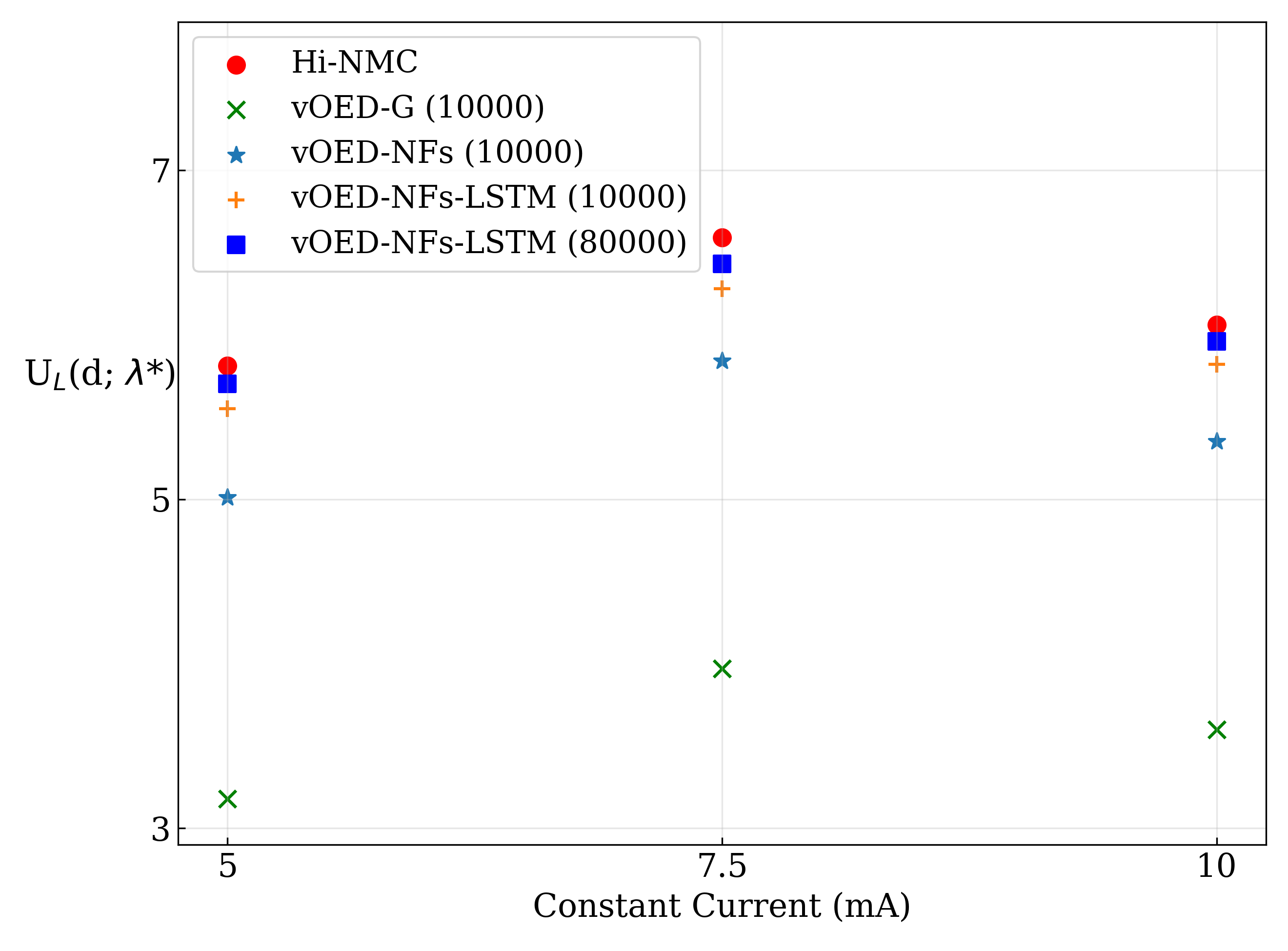}
    \caption{Case 3. Lower bound estimates for all designs using different  methods. Hi-NMC is the high-quality reference EIG estimate.} 
    \label{fig: Ecoat_lower_bounds}
\end{figure}

\Cref{fig: Ecoat Posterior Comparison} presents the marginal and pairwise posterior distributions obtained from using vOED-NFs and vOED-NFs-LSTM lower bound estimates, along with the reference posteriors obtained from SMC. 
We observe that the posteriors obtained from vOED-NFs-LSTM in \cref{fig: Ecoat Posterior Comparison LSTM} offering improved approximation to the SMC reference than those from vOED-NFs in \cref{fig: Ecoat Posterior Comparison noLSTM}. This improvement supports the effectiveness of employing the LSTM summary network for dimension reduction in this example.

\begin{figure}[b!]
    \centering
    \subfloat[vOED-NFs]
    {\includegraphics[width=0.5\textwidth]
    {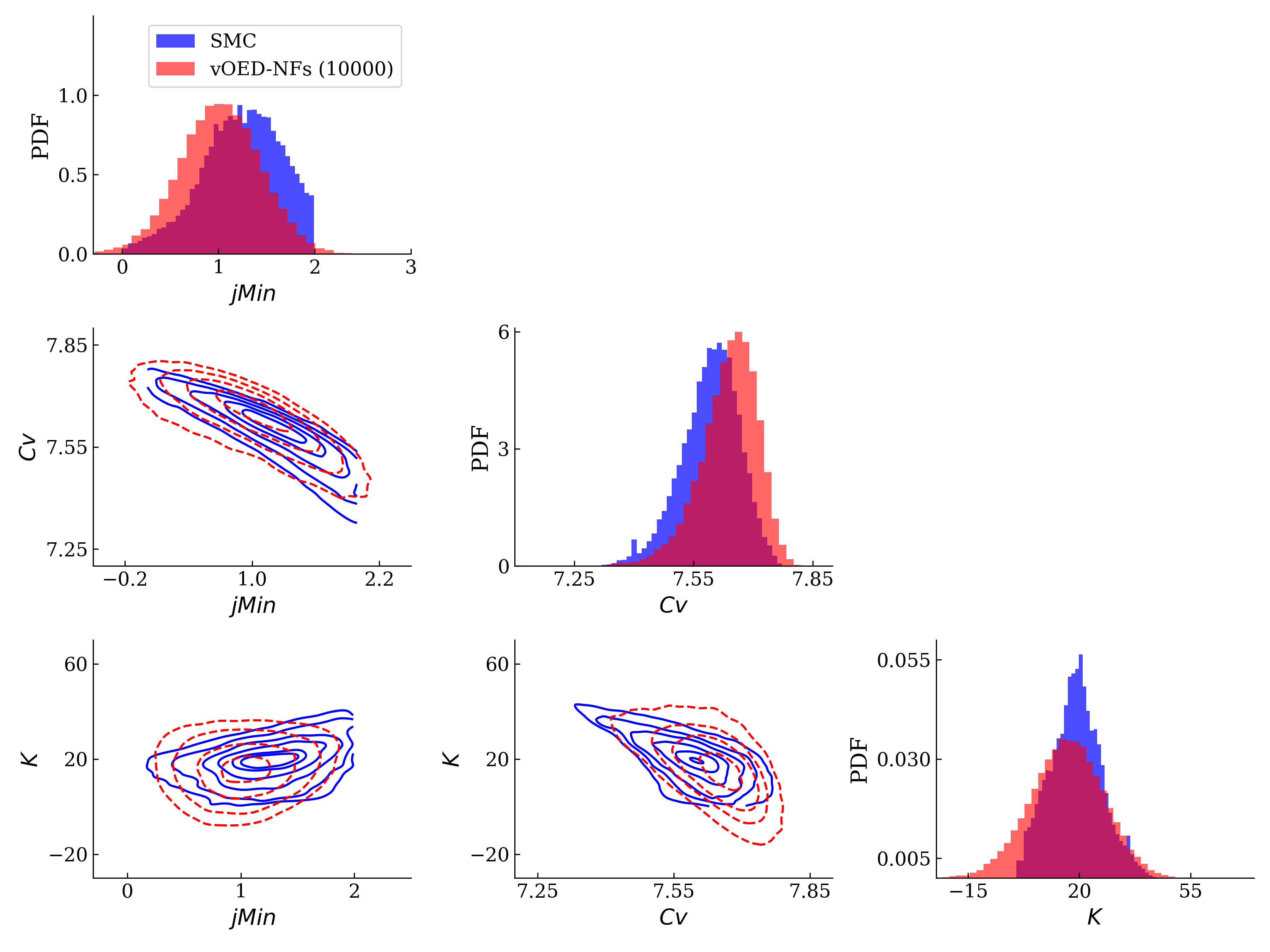}\label{fig: Ecoat Posterior Comparison noLSTM}} 
    \subfloat[vOED-NFs-LSTM]{
    \includegraphics[width=0.5\textwidth]{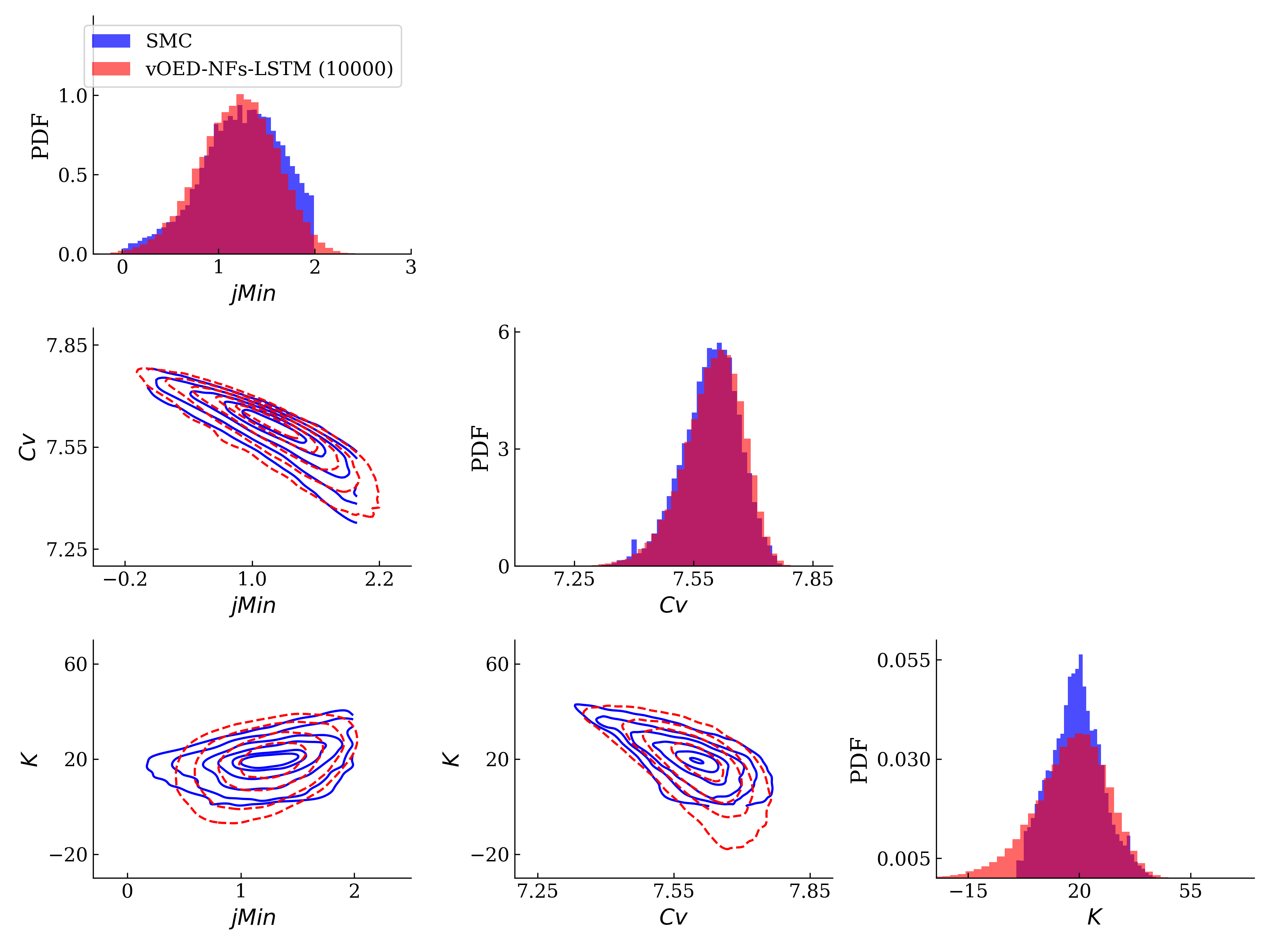}\label{fig: Ecoat Posterior Comparison LSTM}}
    \caption{Case 3. Marginal and pairwise joint posterior distributions obtained using SMC (high-quality reference) and vOED-NFs-LSTM. }
    \label{fig: Ecoat Posterior Comparison}
\end{figure} 

\subsection{Case 4: aphid population}
\label{subsection: case4}

The last case involves a stochastic model depicting the growth of aphid population \cite{Matis07}, which results in an implicit likelihood scenario where samples may be drawn from the likelihood but probability density cannot be evaluated. Denote $M(t)$ and $C(t)$ respectively the current and accumulative sizes of the aphid population, 
the probability that a birth or death occurs in a small time period $\delta_t$ is~\cite{Gillespie_10_Aphids}:
\begin{align}
    \mathbb{P}\{M(t+\delta_t) = m+1, C(t + \delta_t) = c+ 1) \,|\, M(t)=m, C(t)=c\} &= \alpha m \delta_t + o(\delta_t), \\
    \mathbb{P}\{M(t+\delta_t) = m-1, C(t + \delta_t) = c \,|\, M(t)=m, C(t)=c\} &= \beta m c \delta_t + o(\delta_t),
\end{align} 
where $\boldsymbol{\theta}=\{\alpha,\beta\}$ are the birth and death model parameters, respectively. In practice a non-zero $\delta_t$ must be used, leading to a discretized approximation to the continuum limit.
We adopt a prior distribution following~\cite{Gillespie_10_Aphids}:
\begin{align}
\begin{pmatrix}
\alpha \\
\beta
\end{pmatrix} \sim \CN \[ \begin{pmatrix}
0.246 \\
0.000136 
\end{pmatrix}, 
\begin{pmatrix}
0.0079^2 & 5.8 \times 10^{-8} \\
5.8 \times 10^{-8} & 0.00002^2
\end{pmatrix} \],
\end{align}
and an initial condition of $M(0)=C(0)=28$. 
In each experiment, only $M(t)$ is observed at a particular observation time $t$ and $C(t)$ is not observed. An exact likelihood evaluation of $\mathbb{P}(M(t)=m \,|\, \boldsymbol{\theta})$ thus requires the summation of probabilities along all paths that end at $M(t)=m$ and then marginalizing out $C(t)$; this is generally intractable to compute. 

The OED problem is then to determine the optimal $k$ measurement times, $\mathbf{d} = \{t_1, t_2, ..., t_k\}$, $t_k \in [0, 50]$. Following~\cite{Ziqiao_20_LBKLD}, we solve the design optimization for $k = 1$ and 2 (i.e., designing for respectively 1 and 2 measurement times) through grid search, while for $k = 3$ and 4 we use the simultaneous perturbation stochastic approximation (SPSA) algorithm~\cite{Spall1998}. \Cref{table: implicit comparison} presents the design results obtained using vOED-NFs and the LB-KLD method that was used in \cite{Ziqiao_20_LBKLD}. LB-KLD constructs another lower bound to EIG based on the entropy power inequality and entropy’s concavity property, and adopts the nearest neighbor based entropy estimator within the lower bound estimator. Note that LB-KLD does not involve tightening the lower bound and therefore is not a variational lower bound, but instead directly estimates the lower bound, here using $3 \times 10^4$ samples. For vOED-NFs, we set $N_{\text{opt}} = 2 \times 10^4$ and $N= 10^4$ such that the total number of samples is the same. 
Implementation details can be found in \ref{subsection: case4_hyperparam}.

\begin{table}[htb]
\centering
\captionsetup{width=0.9\linewidth}
 \begin{tabular}{||c |c |c| c| c|} 
 \hline
\;\;$k$\;\; & \;\; Method \;\; & \;\; $\mathbf{d}^*$ \;\; & \;\; $ \widehat{U}_{\text{LB-KLD}}(\mathbf{d}^*)$  
\;\; & \;\;  
$\widehat{U}_L(\mathbf{d}^*;\boldsymbol{\lambda^*})$ 
\;\; \\[.5ex] 
 \hline
 \hline
1 & LB-KLD & (21) & 1.18 (0.012) & 1.22 (0.003) \\
 & vOED-NFs & (21) & 1.18 (0.012) & 1.22 (0.003) \\
\hline
2 & LB-KLD & (17, 28) & 1.86 (0.021) & 1.86 (0.018) \\
 & vOED-NFs & (17, 27) & 1.88 (0.008) & 1.89 (0.004) \\
\hline
3 & LB-KLD & (15.7, 22.7, 32.0)  & 2.00 (0.019) & 2.08 (0.004) \\
 & vOED-NFs & (14.6, 20.7, 28.6) & 2.00 (0.018) & \;\; 2.10 (0.005) \;\;\\
\hline
4 & LB-KLD & \;\;(13.8, 19.1, 24.5, 30.6)\;\; & 2.05 (0.017) & 2.20 (0.006)  \\
 & vOED-NFs & \;\;(13.7, 19.2, 24.8, 32.7)\;\; & 2.06 (0.014) & 2.19 (0.006) \\
\hline
\end{tabular}
\caption{Case 4. The third column shows the optimal design $\mathbf{d}^*$  found by LB-KLD and vOED-NFs. The fourth column shows the mean of 10 $\widehat{U}_{\text{LB-KLD}}$ estimates with different random seeds (each with $3 \times 10^4$ samples for Monte Carlo estimation) at the optimal designs, and ($\cdot$) indicates the standard deviation originates from different random partitions of the same 30000 samples used for entropy estimation. The last column shows the mean of 10 $\widehat{U}_{L}$ estimates (each with $N_{\text{opt}} = 2 \times 10^4$ and $N = 10^4$) at the optimal designs but different $\boldsymbol{\lambda^*}$. Here ($\cdot$) indicates the standard deviation due to 10 random seeds for different initialization when optimizing $\boldsymbol{\lambda^*}$, and evaluation is conducted on the same $N = 10^4$ samples.}
\label{table: implicit comparison}
\end{table}

\Cref{table: implicit comparison} indicates that both methods generally propose comparable optimal designs for various $k$ values and similar lower bound estimates. Notably at $k = 3$ and 4, vOED-NFs produce tighter bound estimators than LB-KLD. Samples from the approximate posterior in vOED-NFs for a simulated $\mathbf{y}$ at $k=4$ are illustrated in \cref{fig: Aplid Posterior Comparison}, which agree well with reference posterior samples obtained from approximate Bayesian computation (ABC)~\cite{Marin_11_ABC}. Note that posterior information is only available owing to vOED-NFs's use of $q$ to approximate the posterior density; in contrast, other lower bound approaches not based on posterior approximation, for example LB-KLD, do not offer any posterior information. 

\begin{figure}[htb]
    \centering
    \includegraphics[width=0.65\textwidth]
    {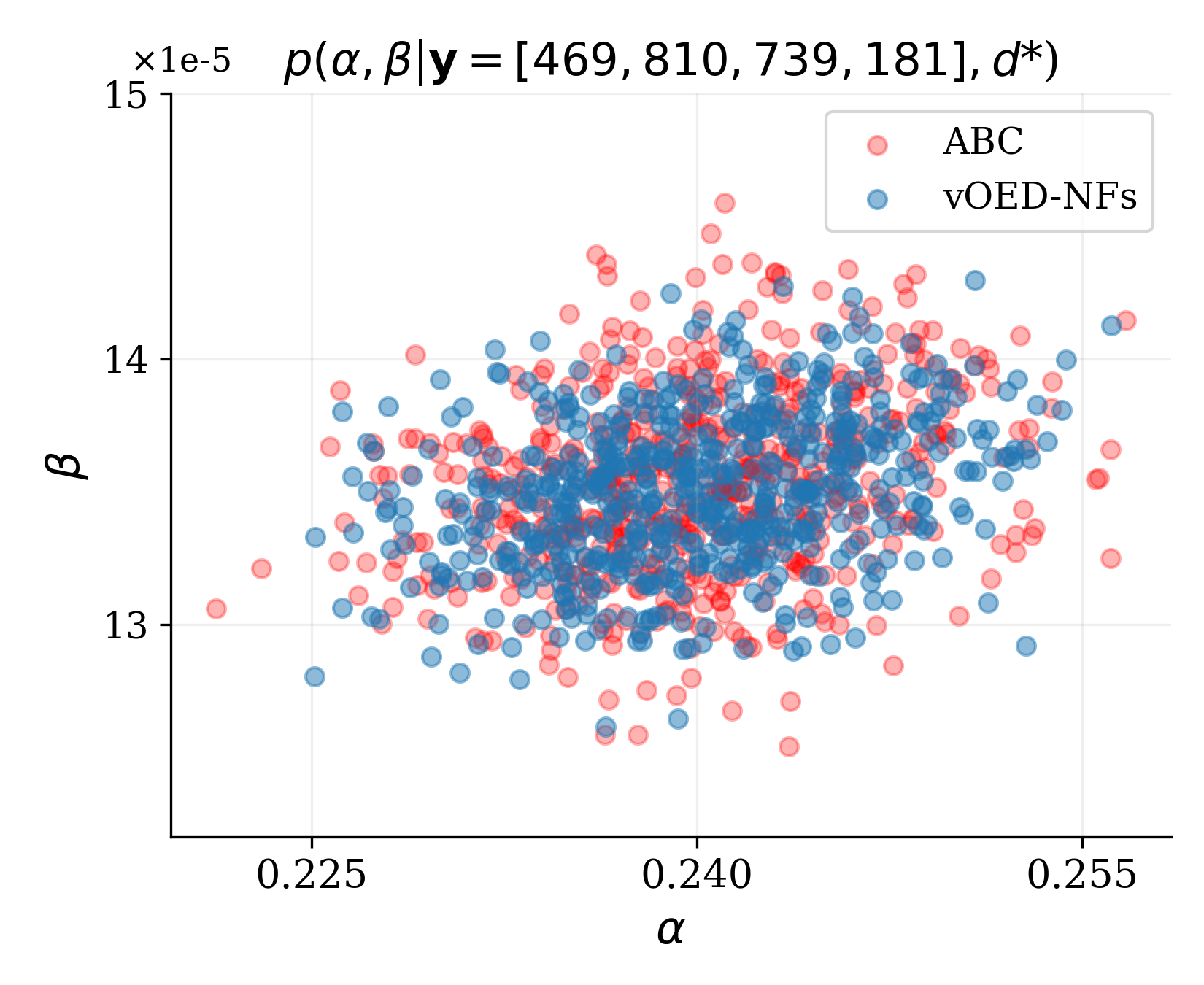}
    \caption{Case 4. Comparison of joint  posteriors samples from ABC and vOED-NFs at the optimal design.}
    \label{fig: Aplid Posterior Comparison}
\end{figure} 

\section{Conclusions}
\label{s: Conclusions}

This paper introduced vOED-NFs, a method to use normalizing flows  to represent variational distributions in the context of Bayesian OED. When the expected utility of OED is chosen to be the EIG in the model parameters, the Barber--Agakov lower bound may be used to estimate the EIG and the bound tightened by optimizing its variational parameters. We presented Monte Carlo estimators to both lower and upper bound versions along with their gradient expressions. We then detailed the use of NFs, in particular with cINN architecture involving a composition of coupling layers and together with a summary network for dimension reduction, to approximate the posterior or marginal likelihood distributions. 

We validated vOED-NFs against established methods in two benchmark problems, and demonstrated vOED-NFs on a PDE-governed application of cathodic electrophoretic deposition and stochastic modeling of aphid population with an implicit likelihood. The findings suggested that 4--5 compositions of the coupling layers were adequate to achieve a lower bias compared to previous approaches. Furthermore, we illustrated that vOED-NFs produced approximate posteriors that matched very well with the true posteriors, able to capture non-Gaussian and multi-modal features effectively.

A limitation of vOED-NFs is the lack of analytical results connecting  bound estimator quality to design optimization. For example, even if the lower bounds are not tight but the bound gap is consistent across $\mathcal{D}$, the design maximizer may still be similar to the true optimal design. Another limitation is that, to retain good accuracy under high-dimensional $\boldsymbol{\theta}$, complex NFs architectures with high-dimensional $\boldsymbol{\lambda}$ may be needed, making the optimization problem challenging. An associated potential difficulty arises from the general susceptibility of neural network training to local minima.
Furthermore,  although we have illustrated greater sampling efficiency of vOED-NFs compared to NMC, it still require a non-trivial number of forward model runs  to tighten the lower bound and achieve high-quality posteriors; further improving sampling efficiency will be important.
Future research exploring efficient and scalable transport map architectures in vOED, such as with rectification operator~\cite{Baptista_23_Representation} and
probability flow ODEs~\cite{Song_21_Maximum}, will be useful. 
The adoption of vOED-NFs into other OED structures, such as  goal-oriented OED \cite{Zhong_24_GOOED} and sequential OED \cite{Shen_23_SOED}, will also be important. 

More broadly, OED approaches that rely on EIG bounds can only be used when the expected utility is the EIG (possibly with other terms that do not depend on the posterior) and cannot be applied to other choices of design criteria. Moreover, the bounds (i.e., \cref{e:UL,e:UU}) cannot be computed in closed form, but only estimated through, for example, Monte Carlo sampling (i.e., \cref{e:Ud_lower_MC,e:Ud_upper_MC}). Consequently, the bound \emph{estimators} can no longer provide inequality guarantees due to the sampling variance. Thus, as a candidate for future exploration, there is a great need for sampling efficiency for bound-based strategies,  to keep the bound estimator variance lower than the bias (i.e., the bound gap).

\section*{Acknowledgement}
\label{s: Acknowledgement}

This work was supported at the University of Michigan by Ford Motor Company under the grant ``Hybrid Physics-Machine Learning Models for Electrodeposition''.
This research was supported in part through computational resources and services provided by Advanced Research Computing at the University of Michigan, Ann Arbor.

\clearpage
\bibliographystyle{elsarticle-num}
\bibliography{main}

\clearpage
\appendix

\section{Proof for \cref{prop:ordering}}
\label{app:prop_proof}

\begin{proof}
Starting from \cref{e:UL_ordering}, we have
\begin{align}
& U_L(\bd;\boldsymbol{\lambda}_1)\leq U_L(\bd;\boldsymbol{\lambda}_2)\nonumber\\
&\iff \mathbb{E}_{\boldsymbol{\Theta},\mathbf{Y}|\mathbf{d}} 
    \[ \ln \frac{q(\boldsymbol{\theta}| \mathbf{y};\boldsymbol{\lambda}_1)}{p(\boldsymbol{\theta} )} \] \leq \mathbb{E}_{\boldsymbol{\Theta},\mathbf{Y}|\mathbf{d}} 
    \[ \ln \frac{q(\boldsymbol{\theta}| \mathbf{y};\boldsymbol{\lambda}_2)}{p(\boldsymbol{\theta} )} \] \nonumber\\
&\iff \mathbb{E}_{\boldsymbol{\Theta},\mathbf{Y}|\mathbf{d}} 
    \[ \ln {q(\boldsymbol{\theta}| \mathbf{y};\boldsymbol{\lambda}_1)}-\ln {p(\boldsymbol{\theta} )} \] \leq \mathbb{E}_{\boldsymbol{\Theta},\mathbf{Y}|\mathbf{d}} 
    \[ \ln {q(\boldsymbol{\theta}| \mathbf{y};\boldsymbol{\lambda}_2)}-\ln {p(\boldsymbol{\theta} )} \] \nonumber\\
&\iff \mathbb{E}_{\boldsymbol{\Theta},\mathbf{Y}|\mathbf{d}} 
    \[ \ln {q(\boldsymbol{\theta}| \mathbf{y};\boldsymbol{\lambda}_1)}-\ln p(\btheta|\by,\bd) \] \leq \mathbb{E}_{\boldsymbol{\Theta},\mathbf{Y}|\mathbf{d}} 
    \[ \ln {q(\boldsymbol{\theta}| \mathbf{y};\boldsymbol{\lambda}_2)}-\ln p(\btheta|\by,\bd) \] \nonumber\\
&\iff \mathbb{E}_{\boldsymbol{\Theta},\mathbf{Y}|\mathbf{d}} 
    \[ \ln \frac{q(\boldsymbol{\theta}| \mathbf{y};\boldsymbol{\lambda}_1)}{p(\btheta|\by,\bd)} \] \leq \mathbb{E}_{\boldsymbol{\Theta},\mathbf{Y}|\mathbf{d}} 
    \[ \ln \frac{q(\boldsymbol{\theta}| \mathbf{y};\boldsymbol{\lambda}_2)}{p(\btheta|\by,\bd)} \] \nonumber\\
&\iff -\mathbb{E}_{\boldsymbol{\mathbf{Y}|\mathbf{d}}} \mathbb{E}_{\boldsymbol{\Theta}|\mathbf{Y},\mathbf{d}} 
    \[ \ln \frac{p(\btheta|\by,\bd)}{q(\boldsymbol{\theta}| \mathbf{y};\boldsymbol{\lambda}_1)} \] \leq -\mathbb{E}_{\boldsymbol{\mathbf{Y}|\mathbf{d}}}\mathbb{E}_{\boldsymbol{\Theta}|\mathbf{d}} 
    \[ \ln \frac{p(\btheta|\by,\bd)}{q(\boldsymbol{\theta}| \mathbf{y};\boldsymbol{\lambda}_2)} \] \nonumber\\
&\iff \EE_{\bY|\bd}\left[\DKL(p_{\btheta|\by,\bd}\,||\,q_{\btheta|\by;\boldsymbol{\lambda}_1})\right] \geq \EE_{\bY|\bd}\left[\DKL(p_{\btheta|\by,\bd}\,||\,q_{\btheta|\by;\boldsymbol{\lambda}_2})\right],\nonumber
\end{align}
ending at \cref{e:EKL_ordering}. The fourth line is from adding $\mathbb{E}_{\boldsymbol{\Theta},\mathbf{Y}|\mathbf{d}}\[\ln {p(\boldsymbol{\theta} )}-\ln p(\btheta|\by,\bd)\]$, which is a constant, on both sides to the preceeding line. 
\end{proof}

\section{Additional Discussions for Invertible Neural Networks}
By incorporating the data $\mathbf{y}$ into the coupling layer, the vOED-NFs can then produce a posterior distribution conditioned on that data. Correspondingly,  \cref{eq:inn_f1,eq:inn_f2} become
\begin{align}
    f_1(\widetilde{\boldsymbol{\Theta}}|\mathbf{y}) &= \begin{bmatrix}
\widetilde{\boldsymbol{\Theta}}_1 \\
\mathbf{Z}_2 = \widetilde{\boldsymbol{\Theta}}_2 \odot \exp(s_1(\widetilde{\boldsymbol{\Theta}}_1, \mathbf{y})) + t_1(\widetilde{\boldsymbol{\Theta}}_1, \mathbf{y})
\end{bmatrix},  \label{eq:inn_f1_with_y}\\
f_2(f_1(\widetilde{\boldsymbol{\Theta}}|\mathbf{y})) &= \begin{bmatrix}
\mathbf{Z}_1 = \widetilde{\boldsymbol{\Theta}}_1 \odot \exp (s_2(\mathbf{Z}_2), \mathbf{y}) + t_2(\mathbf{Z}_2, \mathbf{y}) \\
\mathbf{Z}_2
\end{bmatrix} .
\label{eq:inn_f2_with_y}
\end{align}
The invertibility is ensured as follows. Once a sample for $\widetilde{\boldsymbol{\Theta}}$ is obtained, the outputs from $s_1(\cdot), s_2(\cdot)$ and $t_1(\cdot), t_2(\cdot)$ are then fixed. The resulting transformation for $\mathbf{Z}_2 = \widetilde{\boldsymbol{\Theta}}_2 \odot \exp(s_1(\widetilde{\boldsymbol{\Theta}}_1, \mathbf{y})) + t_1(\widetilde{\boldsymbol{\Theta}}_1, \mathbf{y})$ is then a linear function with respect to $\widetilde{\boldsymbol{\Theta}}_2$ and therefore invertible.

\section{Case 1}
\label{section: case1_appendix}

\subsection{Hyperparameters}
\label{subsection: case1_hyperparam}
The hyperparameters for \cref{subsection: case1} are given in \cref{tab: hyperparam vOED-NFs,tab: hyperparam vOED-G}.
The number of vOED-NFs transformation is also compared across $T=\{1, 3, 5, 7\}$.  

\begin{table}[H]
    \centering
    \begin{tabular}{ll}
    \toprule
     Hyperparameter & vOED-G  \\
    \midrule 
     $N_{\text{opt}}$ &  $2\times10^4$ \\
     $N$ & $10^4$   \\
    $N_{\text{batch}}$ & $1000$  \\
     Initial learning rate & $10^{-2}$ \\
     Learning rate decay \;\; & $0.99$ \\
     Network structure  \;\;  & \{32, 32\} \\
     Training epochs & $301$   \\
     Activation  & ReLU \\
    \bottomrule
    \end{tabular}
    \caption{Case 1. Hyperparameters for vOED-G.}
    \label{tab: hyperparam vOED-G}
\end{table}

\begin{table}[H]
    \centering
    \begin{tabular}{ll}
    \toprule
     Hyperparameter & vOED-NFs  \\
    \midrule 
     $N_{\text{opt}}$ & $5000$ / $10^4$ / $2\times10^4$ / $5\times 10^4$  \\
     $N$ & $10^4$   \\
    $N_{\text{batch}}$ & $250$ / $500$ / $1000$ / $2500$ \\
     Initial learning rate & $5\times 10^{-3}$ / $5\times 10^{-3}$ / $10^{-2}$ / $10^{-2}$\\
     Learning rate decay & $0.99$ \\
     Network structure  (s \& t) \;\;  & \{32, 32\} \\
     Training epochs & $301$   \\
     Activation  & ELU \\
    \bottomrule
    \end{tabular}
    \caption{Case 1. Hyperparameters for vOED-NFs.}
    \label{tab: hyperparam vOED-NFs}
\end{table}

\subsection{EIG estimated on training sample}
\label{subsection: case1_EIG_Training}
\Cref{fig: Assess cost and flexibility of different transformation (training)} shows the EIG evaluated using the same samples it used for optimization (i.e., it is `testing' on the same samples that it used for `training'). We see that in this case, smaller $N_{\text{opt}}$ can actually yield higher EIG estimates than Hi-NMC that uses $N_{\text{out}} = N_{\text{in}} = 2\times10^4$ samples, indicating an overfitting phenomenon where $\widehat{U}_L(d; \boldsymbol{\lambda}^*)$ could overestimate the EIG. Though in this case it does alter the location of the optimal design, the observation suggests the importance of evaluating EIG on a separate sample set than those used for optimization in order to reduce the bias in estimating the lower bound, especially when $N_{\text{opt}}$ is  small.

\begin{figure}[H]
 \centering
  \includegraphics[width=1.0\textwidth]{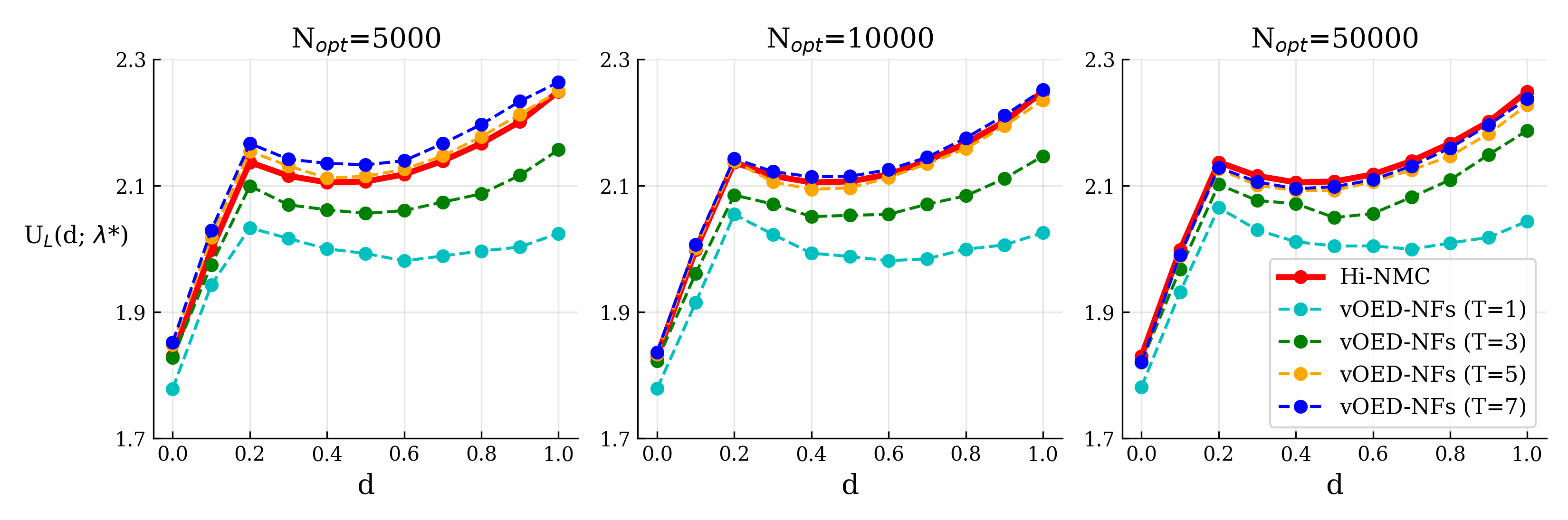}%
  \caption{Case 1. EIG under different number of transformations T and optimization sample sizes $N_{\text{opt}}$, evaluated using the same samples that it used for performing optimization.}
  \label{fig: Assess cost and flexibility of different transformation (training)}
\end{figure}

\section{Case 2}
\label{section: case2_appendix}

\subsection{Hyperparameters}
\label{subsection: case2_hyperparam}

The hyperparameters for \cref{subsection: case2} are given in \cref{tab:hyper_Toy2_vOED-NFs,tab:hyper_Toy2_vOED-GG}.

\begin{table}[H]
    \centering
    \begin{tabular}{ll}
    \toprule
     Hyperparameter & vOED-GG  \\
    \midrule 
     $N_{\text{opt}}$ & $5\times10^4$  \\
     $N$ & $10^4$     \\
     $N_{\text{batch}}$ & $2048$   \\
     Initial learning rate & $5\times 10^{-3}$  \\
     Learning rate decay & $0.99$   \\
     Network structure & \{64, 64\}   \\
     Training epochs & $201$ (\cref{table: regression results}) / $301$ (\cref{fig: regression posterior-AF})  \\
     Activation for  vOED-GG \; & ReLU \\
    \bottomrule
    \end{tabular}
    \caption{Case 2. Hyperparameters for vOED-GG.}
    \label{tab:hyper_Toy2_vOED-GG}
\end{table}

\begin{table}[H]
    \centering
    \begin{tabular}{ll}
    \toprule
     Hyperparam & vOED-NFs  \\
    \midrule 
     $N_{\text{opt}}$ & $5\times10^4$  \\
     $N$ & $10^4$     \\
     $N_{\text{batch}}$ & $2048$ (\cref{table: regression results}) / $1024$ (\cref{fig: regression posterior-NF}) \\
     Initial learning rate & $5\times 10^{-3}$  \\
     Learning rate decay & $0.99$   \\
     Network structure vOED-NFs (s \& t) \;\; & \{32, 32\}   \\
     $T$ (numebr of transformations) $\;\;\;$ & $4$   \\
     Training epochs & $201$ (\cref{table: regression results}) / $301$ (\cref{fig: regression posterior-NF})  \\
     Activation for vOED-NFs \; & ELU \\  
    \bottomrule
    \end{tabular}
    \caption{Case 2. Hyperparameters for vOED-NFs.}
    \label{tab:hyper_Toy2_vOED-NFs}
\end{table}

\section{Case 3}
\label{section: case3_appendix}
\subsection{Forward model}
\label{subsection: case3_Forward_Model}
This section introduces the forward model used in \cref{subsection: case3}.
The model solve for e-coating can be described by three steps. 
First, the electric field within the bath is computed using the conservation of current density. For a constant current e-coating, a Poisson PDE with Robin boundary condition at the interface bath/film and Neumann condition at the anode is solved and described by the following equations:
\begin{align}
\nabla \cdot \mathbf{j}&=0 \label{eq:FluxConservation} \\ 
\mathbf{j}&=\sigma_{\rm{bath}}\nabla \phi \label{eq:CurrentDensity}\\
\left.\phi\right|_{\Gamma}&=R_{\textrm{film}} \,j_n &\textrm{at the interface film-bath}\label{eq:RobinBC}\\
j_n&= j_0& \textrm{at the anode}
\end{align}
where $\mathbf{j}$ is the current density, $j_n=\mathbf{j}\cdot \bf{n}$ is the normal component of the current density, $j_0$ is the prescribed density current at the anode, $\phi$ is the electrical potential, $\sigma_{\rm{bath}}$ is the bath conductivity, $R_{\textrm{film}}$ is the coating film resistance, and $\Gamma$ represents the interface between the coating film and the bath.
Second, the film deposition rate is computed using
\begin{align}
\frac{dh}{dt}=C_v j_n, \label{FilmGrowth}
\end{align}
where $h$ is the film thickness and $C_v$ is the Coulombic efficiency.
Third, the film resistance is found by solving the following equation:
\begin{align}
\frac{dR_{\textrm{film}}}{dt}= \rho(\mathbf{j}) \frac{dh}{dt},\label{ResistanceEquation}
\end{align}
where $ \rho(\mathbf{j})$ is the film resistivity.
The coulombic efficiency $C_v$ is assumed to be constant and the film resistivity $\rho(\mathbf{j})$ is a decreasing function of the current density.

The onset condition is critical for the prediction of the film deposition. Two criteria are used to evaluate the deposition onset. The first one considers a minimal value of the current density $j_{\min}$ that will trigger the deposition:
\begin{align}\label{eq:dhdt_baseline}
\frac{dh}{dt} = C_v\,j_n \textrm{ for } j> j_{\min}.
\end{align}
The second condition is a minimum charge condition $Q_{\min}$ and assumes that the deposition starts when the accumulative charge on the cathode reaches a minimum value as follows:
\begin{align} 
\frac{dh}{dt} = C_v\,j_n \textrm{ for } Q> Q_{\min},
\end{align}
where the electric charge $Q$ is defined as $Q(t)=\int_t j_n dt$. The minimum charge can be expressed as a function of the constant current $j_0$, with $K$ a constant for a given bath and materials:
\begin{align}
Q_{\min}=\frac{K^2}{j_0}.
\end{align}

The truncated normal priors for $\{ C_v, j_{\min}, Q_{\min} \}$ are:
\begin{align}
    &p(C_v) \sim p_{\textit{TN}}(\mu = 7, \; \sigma = 0.5, \; l = 6,\;  u = 8), \\
    &p(j_{\min}) \sim p_{\textit{TN}}(\mu = 1, \; \sigma = 0.5, \; l = 0, \; u = 2), \\
    &p(Q_{\min}) \sim p_{\textit{TN}}(\mu = 50, \sigma = 25, \; l = 0, \; u = 100),
\end{align}
where 
$$p_{\textit{TN}}(x;, \mu, \sigma, l, u) = \frac{1}{\sigma}\frac{\phi(\frac{x-\mu}{\sigma})}{\Phi(\frac{u-x}{\sigma})-\Phi(\frac{l-x}{\sigma})}$$ for $x \in [l, u]$, and equals 0 otherwise. Here $\phi(\cdot)$ denotes the PDF for a standard normal distribution and $\Phi (\cdot)$ is its cumulative distribution function.

\subsection{Hyperparameters for the 10-dimensional case}
For $\mathbf{y}$ with dimension 10, \cref{tab: hyper_case3_10} lists the hyperparameters used for the lower and upper bound estimators in vOED-NFs. 
\label{subsection: case3_10_upper_hyperparam}

\begin{table}[H]
    \centering
    \begin{tabular}{ll}
    \toprule
     Hyperparam & vOED-NFs  \\
    \midrule 
    $N_{\text{opt}}$ & $10^4$ / $8\times10^4$  \\
    $N$ & $10^4$     \\
    $N_{\text{batch}}$ & $512$ / $2048$   \\
     Initial learning rate & $5\times 10^{-3}$  \\
     Learning rate decay & $0.99$   \\
     Network structure  (s \& t) \; & \{16, 16\}   \\
     $T\;\;\;$ & $5$   \\
     Training epochs & $301$  \\
     Activation for vOED-NFs & ELU     \\
    \bottomrule
    \end{tabular}
    \caption{Case 3. Hyperparameters for the 10-dimensional case, both lower and upper bound estimators.}
    \label{tab: hyper_case3_10}
\end{table}

\subsection{Hyperparameters for the 50-dimensional case}
\label{subsection: case3_hyperparam_50points}
For $\mathbf{y}$ with dimension 50, \cref{tab:hyper_case3_50_vOED-G} lists the hyperparameters used for the lower bound estimators in vOED-G, and 
\cref{tab: hyper_case3_50} lists the hyperparameters used for the lower and upper bound estimators in vOED-NFs, without the LSTM summary network. 

\begin{table}[H]
    \centering
    \begin{tabular}{ll}
    \toprule
     Hyperparameter & vOED-G  \\
    \midrule 
     $N_{\text{opt}}$ & $10^4$  \\
     $N$ & $10^4$     \\
     $N_{\text{batch}}$ & $256$   \\
     Initial learning rate & $1\times 10^{-3}$  \\
     Learning rate decay & $0.99$   \\
     Network structure & \{32, 32\}   \\
     Training epochs & 301  \\
     Activation for  vOED-G \; & ReLU \\
    \bottomrule
    \end{tabular}
    \caption{Case 3. Hyperparameters for vOED-G for the 50-dimensional case, for the lower bound estimator.}
    \label{tab:hyper_case3_50_vOED-G}
\end{table}

\begin{table}[H]
    \centering
    \begin{tabular}{ll}
    \toprule
     Hyperparameter &  vOED-NFs  \\
    \midrule 
     $N_{\text{opt}}$ & $10^4$ / $8\times 10^4$  \\
     $N$ & $10^4$     \\
     $N_{\text{batch}}$ & $512$ / $2048$   \\
     Initial learning rate & $1\times 10^{-3}$/$5\times 10^{-3}$  \\
     Learning rate decay & $0.99$   \\
     Network structure (s \& t)  \; & \{16, 16\}   \\
     $T \;\;\;$ & $5$   \\
     Training epochs & $301$  \\
     Activation for vOED-NFs & ELU     \\
    \bottomrule
    \end{tabular}
    \caption{Case 3. Hyperparameters for vOED-NFs for the 50-dimensional case, for both lower and upper bound estimators without LSTM summary network.}
    \label{tab: hyper_case3_50}
\end{table}

\Cref{tab: hyper_case3_50_LSTM} lists the hyperparameters for vOED-NFs with the LSTM summary network. %
The initial learning rates reported in both tables with or without LSTM empirically yield the highest lower bound estimates among the learning rate values $\{5\times 10^{-4}, 1 \times 10^{-3}, 5\times 10^{-3}, 1\times 10^{-2}\}$. To get a dimension reduced 10-dimensional $\mathbf{y}'$, we first feed $\mathbf{y}$ as input to a bidirectional LSTM network, and then concatenate the resulting context vectors, which is further fed into the final embedding network emanating $\mathbf{y}'$.
\begin{table}[htp]
    \centering
    \begin{tabular}{ll}
    \toprule
     Hyperparameters &  vOED-NFs-LSTM  \\
    \midrule 
     $N_{\text{opt}}$  & $10^4$ / $8\times 10^4$  \\
     $N$ & $10^4$     \\
     $N_{\text{batch}}$ & $512$ / $2048$   \\
     Initial learning rate & $5\times 10^{-3}$/$1\times 10^{-2}$ \\
     Learning rate decay & $0.99$   \\
     Network structure (s \& t)  & \{16, 16\}   \\
     $T \;\;\;$ & $5$   \\
     Training epochs & $301$  \\
     Activation for vOED-NFs-LSTM & ELU   \\
     LSTM-n\_feature & $1$  \\
     LSTM-hidden\_dim & $20$  \\
     LSTM-num\_layers & $1$  \\
     Embedding network structure \;  & {40, 20} \\
     Activation for embedding network  \;\;& ELU \\
    \bottomrule
    \end{tabular}
    \caption{Case 3. Hyperparameters for vOED-NFs for the 50-dimensional case, for lower bound estimator
    with LSTM summary network.}
    \label{tab: hyper_case3_50_LSTM}
\end{table}

\subsection{Posteriors from vOED-NFs with LSTM summary network and $N_{\text{opt}} = 8 \times 10^4$}

\Cref{fig: Ecoat Posterior with more samples and LSTM} plots vOED-NFs posteriors when using LSTM summary network and $N_{\text{opt}} = 8 \times 10^4$. The posteriors match the SMC posteriors (reference) better compared to those trained with smaller $N_{\text{opt}}$ and when $\mathbf{y}$ is higher dimensional (i.e., from \cref{fig: Ecoat Posterior Comparison}).
\begin{figure}[H]
\centering
\includegraphics[width=0.6\textwidth]{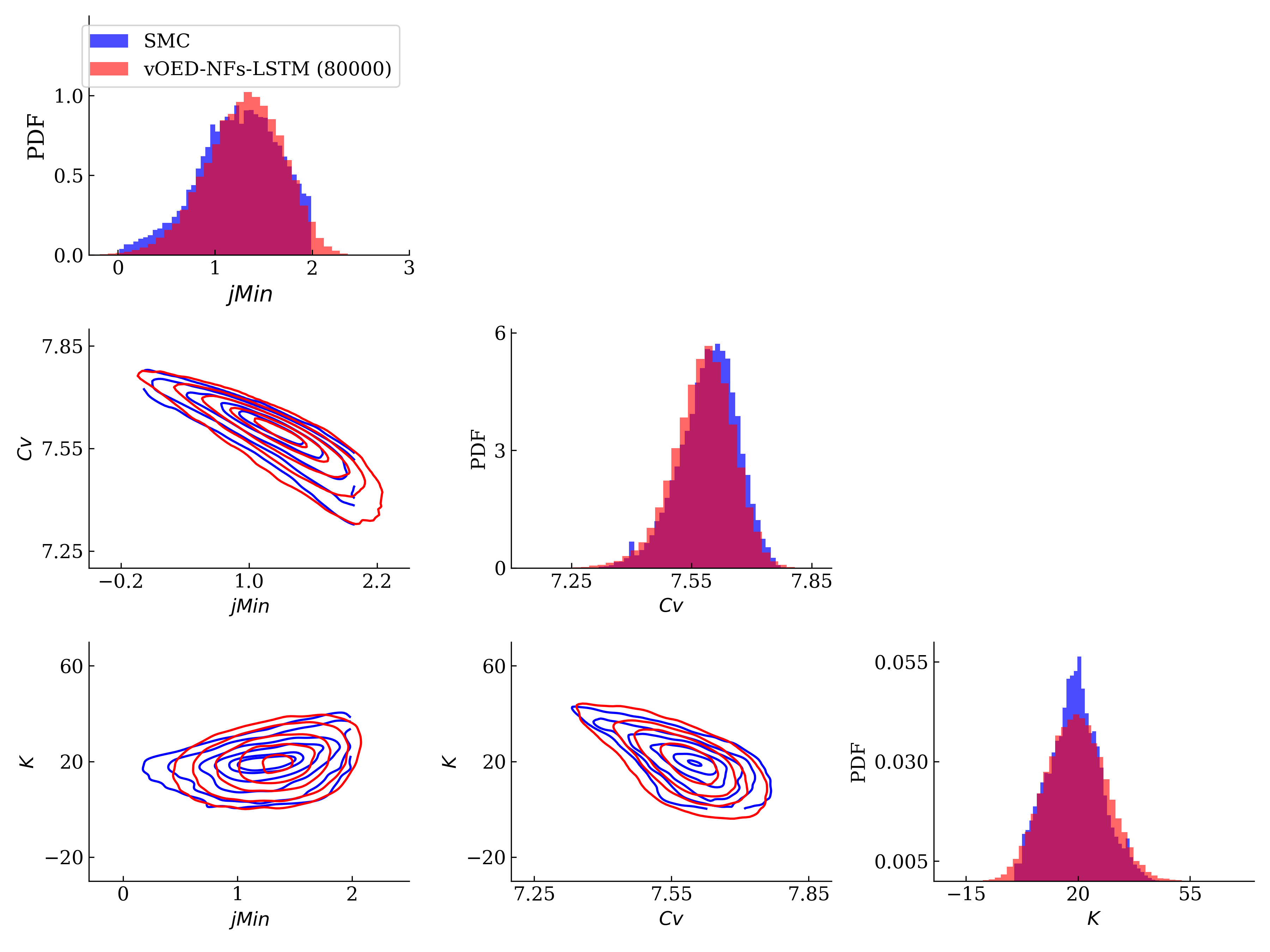} 
\caption{Case 3. Marginal and pairwise joint posterior distributions obtained using SMC (high-quality reference) and vOED-NFs-LSTM with $N_{\text{opt}}=80000$.}
\label{fig: Ecoat Posterior with more samples and LSTM}
\end{figure}

\section{Case 4}
\label{section: case4_appendix}

\subsection{Hyperparameters}
\label{subsection: case4_hyperparam}

The hyperparameters for \cref{subsection: case4} are given in \cref{tab: hyper_Toy4}. Note LB-KLD \cite{Ziqiao_20_LBKLD} is computed using $10^4$ prior samples, each associated with $3$ data samples such that the overall number of forward model runs is $3\times 10^4$. For vOED-NFs, we keep the total number of forward model runs at $3\times 10^4$ ($N_{\text{opt}} = 2\times 10^4$ and $N = 10^4$).

\begin{table}[H]
    \centering
    \begin{tabular}{ll}
    \toprule
     Hyperparameters & vOED-NFs  \\
    \midrule 
     $N_{\text{opt}}$ & $2 \times 10^4$  \\
     $N$ & $10^4$    \\
     $N_{\text{batch}}$ & $2048$ \\
     Initial learning rate  & $1\times 10^{-2}$   \\
     Learning rate  decay & $0.99$  \\
     Network structure (s \& t) \;\;  & \{16, 16\} \;   \\
     $T \;\;\;$ & $4$   \\
     Training epochs & $51$   \\
     Activation for vOED-NFs & ELU  \\
    \bottomrule
    \end{tabular}
    \caption{Case 4. Hyperparameters for vOED-NFs.}
    \label{tab: hyper_Toy4}
\end{table}

\section{Training Time for EIG Estimation}
\label{section: training_time}

This section provides a sense of the computational cost in terms of ``running time'' for EIG evaluation at a specific design. We note that these figures are  dependent on the implementation, such as which packages are being used, which training hyperparameters are adopted, whether code is optimized or parallelized, etc. Therefore, we suggest it to be used for awareness only, rather than direct comparisons. All numerical comparisons are run on the University of Michigan Great Lakes Slurm HPC Cluster nodes, with each node equipped with a single Nvidia Tesla A40 or V100 GPU.

For case 1 in \cref{subsection: case1},
\cref{fig:Toy1_a}, 
the EIG evaluation time at a specific design for approaches (a)--(e)  are approximately: 
(a) (Hi-NMC) 57 seconds, (b) (Lo-NMC) less than 0.1 seconds,   (c) (Re-NMC) 16.5 seconds, (d) (vOED-G) 43 seconds, and (e) 1 minute 38 seconds. For this case, the forward model run is extremely fast and thus the computational cost from vOED is mainly contributed by the optimization process.

For case 2 in \cref{subsection: case2}, \cref{table: regression results}, the improvement in EIG and posterior approximation of vOED-NFs over vOED-GG comes at the expense of increased training time, with vOED-GG requiring approximately 1 minute and 39 seconds per EIG evaluation, while vOED-NFs takes around 5 minutes and 56 seconds under the selected hyperparameters.

For case 3 in \cref{subsection: case3},
\cref{fig: 10-dim result}, 
the training times for (a) vsOED-NFs (upper), $N_{\text{opt}} = 10^4$, (b) vsOED-NFs (upper), $N_{\text{opt}} = 8 \times 10^4$, (c) vsOED-NFs (lower), $N_{\text{opt}} = 10^4$, (d) vsOED-NFs (lower), $N_{\text{opt}} = 8 \times 10^4$ are respectively 2 minutes 32 seconds, 5 minutes 59 seconds, 3 minutes 50 seconds, and 6 minutes 48 seconds under the selected hyperparameters. Whereas for the Hi-NMC estimator which requires $10^5$ forward model runs in total, each run involves solving a PDE and takes approximately 1.5 seconds. Therefore, the additional time induced by training NFs is trivial compared to the overall computational cost of running the forward models for Hi-NMC.

For case 3, \cref{fig: Ecoat Posterior Comparison}, with LSTM incorporated via summary network for evaluating the EIG lower bounds, the training times for (a) vOED-NFs, $N_{\text{opt}} = 10^4$, (b)  vOED-NFs, $N_{\text{opt}} = 8 \times 10^4$, (c)  vOED-NFs-LSTM, $N_{\text{opt}} = 10^4$, (d)  vOED-NFs-LSTM, $N_{\text{opt}} = 8 \times 10^4$ are respectively 4 minutes 33 seconds, 6 minutes 19 seconds, 5 minutes 13 seconds, and 7 minutes 38 seconds under the selected hyperparameters.

Lastly, for case 4 in \cref{subsection: case4}, 
\cref{fig: Aplid Posterior Comparison}, 
the training time for vOED-NFs is approximately 15 seconds.

\end{document}